\documentclass[10pt,twocolumn,letterpaper]{article}

\usepackage{cvpr}              % To produce the CAMERA-READY version
\usepackage{graphicx}
\usepackage{amsmath}
\usepackage{amssymb}
\usepackage{booktabs}
\usepackage{multirow}
\usepackage{multicol}

\usepackage[pagebackref,breaklinks,colorlinks]{hyperref}

\usepackage[capitalize]{cleveref}
\crefname{section}{Sec.}{Secs.}
\Crefname{section}{Section}{Sections}
\Crefname{table}{Table}{Tables}
\crefname{table}{Tab.}{Tabs.}

\def\cvprPaperID{72} % *** Enter the CVPR Paper ID here
\def\confName{CVPR}
\def\confYear{2022}

\begin{document}

% \title{Residual Local Feature Network for Efficient Image Super-Resolution}
\title{Residual Local Feature  Network for Efficient Super-Resolution}

\author{Fangyuan Kong\thanks{Equal contribution} \quad Mingxi Li$^{*}$ \quad Songwei Liu$^{*}$ \quad Ding Liu \quad Jingwen He \\
\quad Yang Bai \quad Fangmin Chen \quad Lean Fu\\ %\thanks{Corresponding author}
ByteDance Inc \\
{\tt\small \{kongfangyuan, limingqi.01, liusongwei.zju, liuding, hejingwen.2021\}@bytedance.com}\\
% {\tt\small \{baiyang.87, chengshu.2019, fulean\}@bytedance.com}
% }
{\tt\small baiyang.87@bytedance.com \quad cfangmin@gmail.com \quad fulean@bytedance.com}
}
\maketitle

%%%%%%%%% ABSTRACT
\begin{abstract}
    Deep learning based approaches has achieved great performance in single image super-resolution (SISR). 
    % However, most of the existing methods require considerable computation costs, which makes them difficult to be deployed to resource-constrained devices. 
    % \textcolor{red}{In this work, we propose a novel residual local feature network (RLFN) which is lightweight and accurate. ***}. 
    However, recent advances in efficient super-resolution focus on reducing the number of parameters and FLOPs, and they aggregate more powerful features by improving feature utilization through complex layer connection strategies. These structures may not be necessary to achieve higher running speed, which makes them difficult to be deployed to resource-constrained devices. In this work, we propose a novel Residual Local Feature Network (RLFN). The main idea is using three convolutional layers for residual local feature learning to simplify feature aggregation, which achieves a good trade-off between model performance and inference time.
    Moreover, we revisit the popular contrastive loss and observe that the selection of intermediate features of its feature extractor has great influence on the performance. Besides, we propose a novel multi-stage warm-start training strategy. In each stage, the pre-trained weights from previous stages are utilized to improve the model performance. Combined with the improved contrastive loss and training strategy, 
    % the proposed RLFN outperforms all the state-of-the-art methods in terms of both PSNR and SSIM for $\times4$ SR task with higher inference speed. 
    the proposed RLFN outperforms all the state-of-the-art efficient image SR models in terms of runtime while maintaining both PSNR and SSIM for SR. 
    % with scale factor of 4.
    In addition, we won the first place in the runtime track of the NTIRE 2022 efficient super-resolution challenge. 
    Code will be available at \href{https://github.com/fyan111/RLFN}{https://github.com/fyan111/RLFN}.
    % Code will be available upon the acceptance of this paper.
    
\end{abstract}

\section{Introduction} \label{sec:intro}
SISR aims to reconstructed a high-resolution image from a low-resolution image. It is a fundamental low-level vision task and has a wide range of applications\cite{FSRNet, DepthMapSR, HyperspectralSR}. Currently, deep learning based approaches\cite{SRCNN, FSRCNN, ESPCN, EDSR, RCAN, CARN, IMDN, RFDN, RFANet, ECBSR, HAN, DFSA, SwinIR, IPT, NLSA} have achieved great success and continuously improved the quality of reconstructed images. However, most of these advanced works require considerable computation costs, which makes them difficult to be deployed on resource-constrained devices for real-world applications. Therefore, it is essential to improve the efficiency of SISR models and design lightweight models that can achieve good trade-offs between image quality and inference time.

\begin{figure}[t]  
\vspace{-1.5em}
	\centering
	\includegraphics[width=\linewidth,scale=1.00]{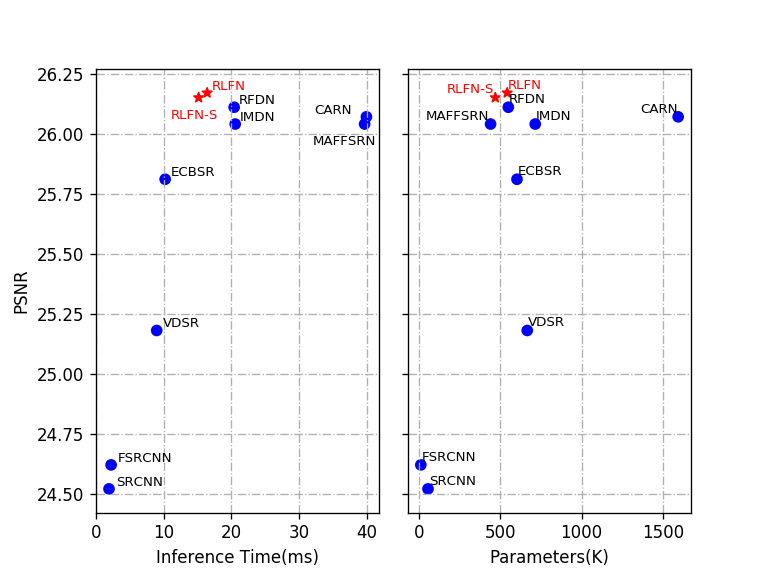}
% 	\caption{ Performance and Inference Time, Parameters comparison between our RLFN and other state-of- the-art lightweight networks on Urban100 dataset for scale factor of 4.}
    \caption{Illustration of PSNR, inference time and parameter numbers of different SISR models on the Urban100 dataset for 4x SR.}
	\label{fig:comparison}
\vspace{-1.5em}
\end{figure}

Many prior arts~\cite{SRCNN, FSRCNN, ESPCN, EDSR, RCAN, CARN, DRCN, DRRN, IMDN, RFDN, ECBSR, AIM2020, MAI2021SR} have been proposed to develop efficient image super-resolution models. Most of these efficient models try to reduce model parameters or FLOPs. To reduce model parameters, recursive networks with weight sharing strategy are usually adopted \cite{DRCN, DRRN}. However, such models do not essentially decrease the number of operations and inference time because the complex graph topology is not reduced. To reduce FLOPs of SR models, it is common to employ operations like depth-wise convolutions, feature splitting and shuffling \cite{CARN, IDN, IMDN, RFDN}, which are without the guarantee of improving computational efficiency.
Some recent studies\cite{AIM2020, ECBSR} have shown that fewer parameters and FLOPs do not always lead to better model efficiency, especially runtime, which is generally the most important factor for practical applications. Parameters and FLOPs are widely used for theoretical analysis but are only proxies for the actual inference time on physical devices. Therefore, there is a high demand to develop efficient SR models that have higher inference speed instead of fewer parameters or FLOPs, in order to better meet practical and commercial needs.

To this end, we revisit current state-of-the-art efficient SR model RFDN\cite{RFDN} and attempt to achieve better trade-offs between reconstructed image quality and inference time. First, we rethink the efficiency of several components of the residual feature distillation block proposed by RFDN. We observe that though feature distillation significantly reduce the number of parameters and contribute to the overall performance, it is not hardware friendly enough and limits the inference speed of RFDN. To improve its efficiency, we propose a novel Residual Local Feature Network (RLFN) that can reduce the network fragments and maintain the model capacity. To further boost its performance, we propose to employ the contrastive loss\cite{AECR-Net, CSD}. We notice that the selection of intermediate features of its feature extractor has great influence on the performance. We conduct comprehensive studies on the properties of intermediate features and draw a conclusion that features from shallow layers preserve more accurate details and textures, which are critical for PSNR-oriented models. Based on this, we build an improved feature extractor that effectively extracts edges and details. To accelerates the model convergence and enhance the final SR restoration accuracy, we propose a novel multi-stage warm-start training strategy. Specifically, in each stage, the SR model can enjoy the benefit of pre-trained weight of models from all previous stages. Combined with the improved contrastive loss and the proposed warm-start training strategy, RLFN achieves state-of-the-art performance and maintain good inference speed. Figure \ref{fig:comparison} shows that RLFN has a better trade-off between image quality and inference time than other recent competitors of efficient SR models. 

Our contributions can be summarized as follows:

\begin{itemize}
    \vspace{-0.5em}
    \item[1.] We rethink the efficiency of RFDN and investigate its speed bottleneck. 
    % We propose a novel architecture Residual Local Feature Network, which can maintain the model capacity and accelerate the inference.
    We propose a novel network termed \textit{Residual Local Feature Network}, which successfully enhances the model compactness and accelerates the inference without sacrificing SR restoration accuracy.
    \vspace{-0.8em}
    \item[2.] We analyze intermediate features extracted by the feature extractor of the contrastive loss. 
    % We draw a conclusion that features from shallow layers are critical for PSNR-oriented models. Based on this, we propose a novel feature extractor and it can extract more edge and texture information.
    We observe that features from shallow layers are critical for PSNR-oriented models, which inspires us to propose a novel feature extractor to extract more information of edges and textures.
    \vspace{-0.8em}
    \item[3.] We propose a multi-stage warm-start training strategy. It can utilize the trained weights from previous stages to boost the SR performance.
    \vspace{-0em}
\end{itemize}

\section{Related Work}

\subsection{Efficient Image Super-Resolution}

\begin{figure*}[!htb]
% \vspace{2em}
  \centering
  \includegraphics[width=0.65\linewidth,
  ]{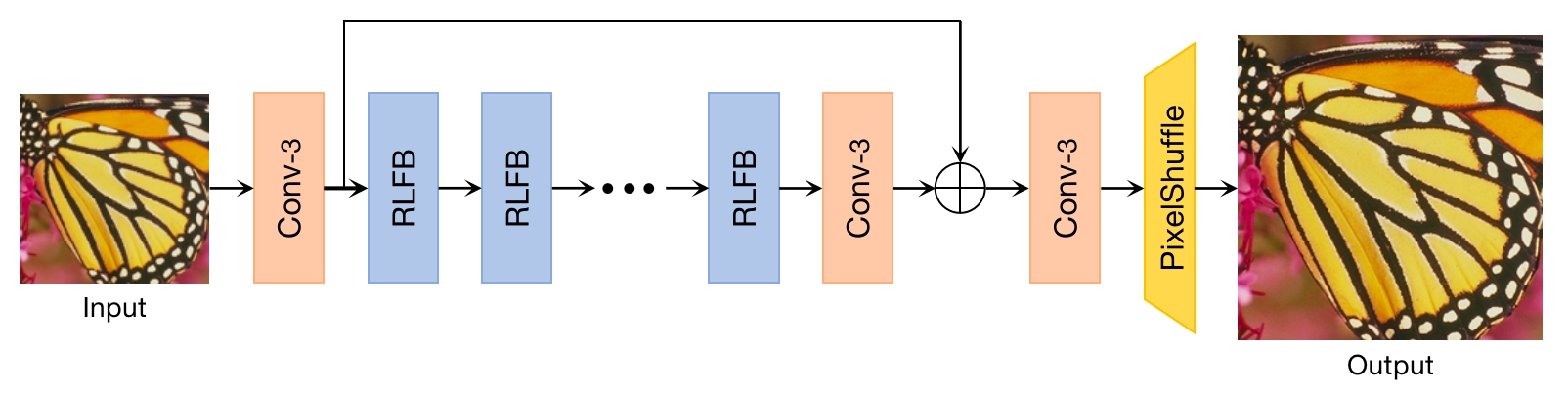}
  \caption{The architecture of residual local feature network.}
  \label{fig:modelarchitecture}
 \vspace{-1em}
\end{figure*}
\begin{figure*}[!htb]
  \centering
  \subfloat[][RFDB]{
    \includegraphics[width=0.3\linewidth, %height=17em,
    ]{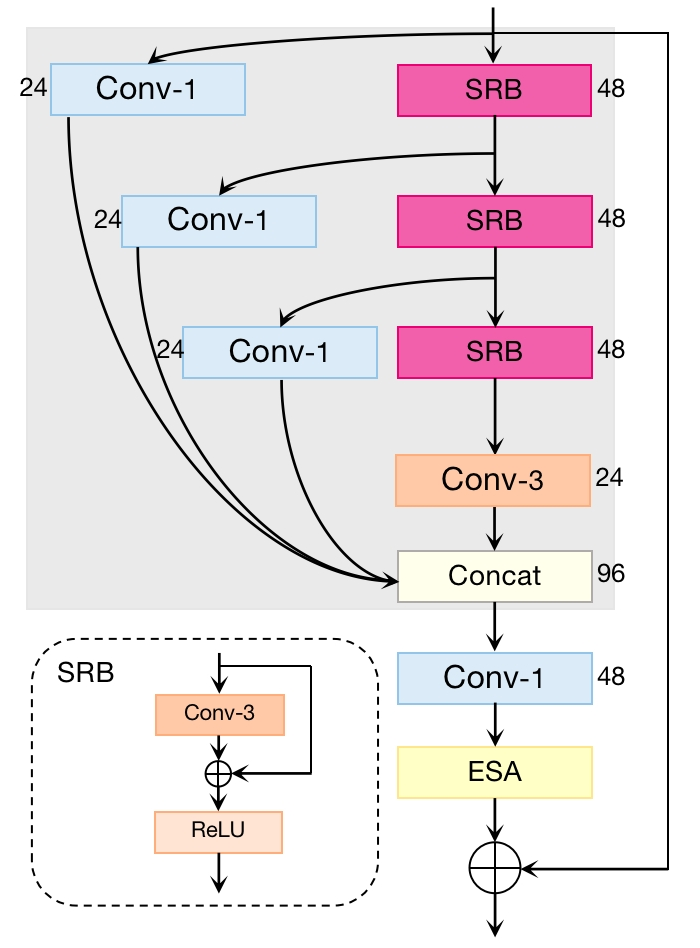}
    \label{fig:RFDBarchitecture}
  }
%   \qquad
%   \subfloat[][]{
%     \includegraphics[width=0.2\linewidth]{./Figures/net_structure/b_TS}
%     \label{fig:Transitionarchitecture}
%   }
%   \qquad
  \subfloat[][RLFB]{
    \includegraphics[width=0.2\linewidth, %height=17em,
    ]{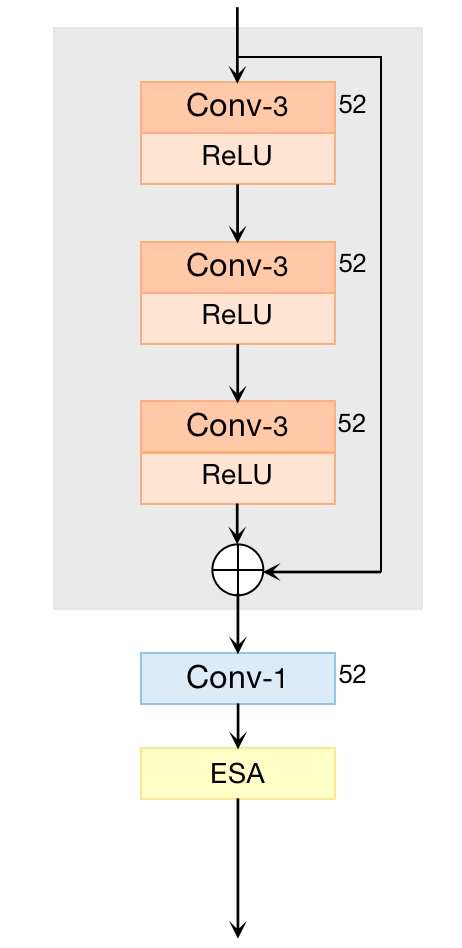}
    \label{fig:RLFBarchitecture}
  }
%   \qquad
  \subfloat[][ESA]{
    \includegraphics[width=0.2\linewidth, %height=17em,
    ]{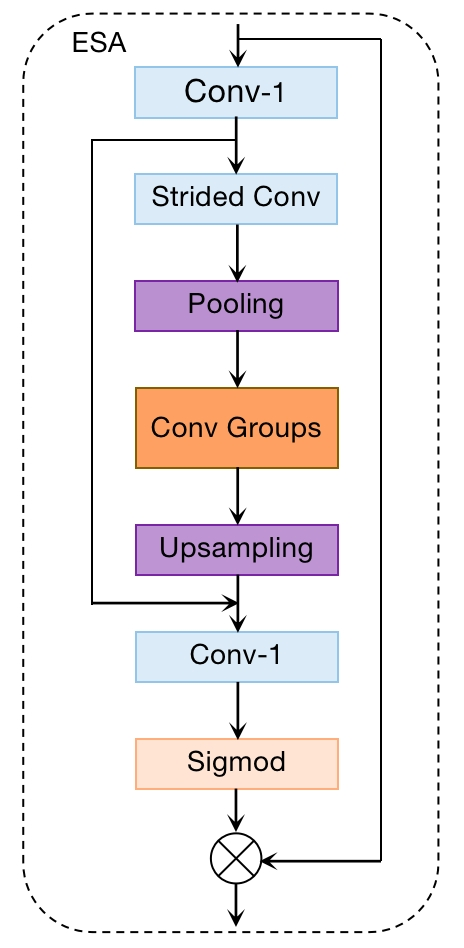}
    \label{fig:ESAarchitecture}
  }
%  \vspace{1em}
  \caption{Conv-1 denotes $1\times1$ convolution and Conv-3 denotes $3\times3$ convolution.  (a) RFDB: residual feature distillation block. (b) RLFB: residual local feature block. (c) ESA: Enhanced Spatial Attention.  In (a) and (b) channel numbers of output feature maps are shown next to each layer.}
\label{fig:blockarchitecture}
\vspace{-1.5em}
\end{figure*}

Achieving real-time SISR on resource-constrained mobile devices has huge business benefits, therefore, we mainly discuss lightweight image SR methods. SCRNN\cite{SRCNN} applied the deep learning algorithm to the SISR field for the first time. It has three layers and uses bicubic interpolation to upscale the image before the net, causing unnecessary computational cost. To address this issue, FSRCNN\cite{FSRCNN} employed the deconvolution layer as the upsampling layer and upscaled the image at the end of net. DRCN\cite{DRCN} introduced a deep recursive convolutional network to reduce the number of parameters. LapSRN\cite{LapSRN} proposed the laplacian pyramid super-resolution block to reconstruct the sub-band residuals of HR images. CARN\cite{CARN} proposed an efficient cascading residual network with group convolution, which obtains comparable results against computationally expensive models. IMDN\cite{IMDN} proposed a lightweight information multi-distillation network by constructing the cascaded information multi-distillation blocks, which extracts hierarchical features step-by-step with the information distillation mechanism (IDM).
%  and won the AIM 2019 constrained image SR challenge\cite{AIM2019}. 
% Although IMDN demonstrates the excellent performance of the information distillation mechanism (IDM), it is not efficient enough and brings some inflexibility in the network design.
% To address above issues, 
RFDN\cite{RFDN} refined the architecture of IMDN and proposed the residual feature distillation network, which replaced IDM with feature distillation connections.
% and won the first place in the AIM 2020 efficient super-resolution challenge\cite{AIM2020}. 
ECBSR\cite{ECBSR} proposed an edge-oriented convolutional block  based on the reparameterization technique\cite{REPVGG}, which can improve the learning ability of the model without increasing the inference time. In order to obtain better results with limited computational effort, the above studies tend to utilize various complex inter-layer connections which affect the inference speed. In this paper we propose a simple network structure with enhanced training strategies to obtain better a trade-off between SR quality and model inference speed.

% \subsection{Structured Pruning}
% The improvement of neural network performance is usually accompanied by the increase of resource requirements such as params and flops, One popular approach for reducing them at test time is neural network pruning, which can be categorized into structured pruning and unstructured pruning. Unstructured pruning\cite{LC}\cite{DP}\cite{LWC} focuses on pruning fine-grained weight of filters, so that leading to unstructured sparsity in models, which cannot be directly accelerated on general computing libraries. In contrast, the pruning target of structured pruning\cite{FP} is entire filter, which could achieve the structured sparsity, so the pruned model could take full advantage of high efficiency Basic Linear Algebra Subprograms (BLAS) libraries to achieve better acceleration. \cite{FP} uses l1-norm to
% select unimportant filters and explores the sensitivity of layers for filter pruning. \cite{liu2017learning} introduces sparsity on the scaling parameters of batch normalization (BN) layers to prune the network. \cite{molchanov2016pruning} proposes a Taylor expansion based pruning criterion to approximate the change in the cost function induced by pruning. These pruning algorithms prune the pretrained model once and retrained once to recover the accuracy. To reduce  dependence on pretrained models and improve model capacity, \cite{SFP} proposed Soft Filter Pruning (SFP) enables the pruned filters to be updated when training the model after pruning, which achieve state-of-the-art results.

\subsection{Train Strategy for PSNR-oriented SISR}
According to machine learning theory, good prediction results come from the combined optimization of architecture, training data, and optimization strategies. Previous works on SISR mainly focused on network architecture optimization\cite{CARN,IMDN,DRCN,EDSR,ECBSR}, while the importance of training strategies that contribute collaboratively to the performances is rarely explored. These SR networks are usually trained by the ADAM optimizer with standard l1 loss for hundreds of epoches. To improve the robustness of training, they usually adopt a smaller learning rate and patch size.
Recent works on image recognition\cite{revisitingresnet} and optical flow estimation\cite{sun2021autoflow} have demonstrated that advanced training strategies can enable older network architectures to match or surpass the performance of novel architectures. Such evidence motivates us to interrogate the training strategies for SR models and unlocking their potential. RFDN\cite{RFDN} demonstrated that both fine-tuning the network with l2 loss and initializing a 4x SR model with pretrained 2x model can effectively improve PSNR. RRCAN\cite{lin2022revisiting} revisited the popular RCAN model and demonstrated that increasing training iterations clearly improves the model performance.

\section{Method}
In this section, we first introduce our proposed RLFN 
% which can achieve a good trade-off between performance and inference time 
in Section~\ref{sec:methods_architecture}. In Section \ref{sec:methods_cl}, we revisit the contrastive loss and analyze several limitations of its feature extractor. Then we propose an improved feature extractor which can provide more stronger guidance during the training process. In Section~\ref{sec:methods_ws}, we describe a novel multi-stage
warm-start training strategy, which can effectively improve the performance of lightweight SR models.

\begin{figure*}[!htb]
%   \vspace{1em}
    \centering
    \begin{subfigure}[H]{0.15\linewidth}
        \centering
        \begin{subfigure}[H]{\linewidth}
            \centering
    		\includegraphics[width=\linewidth]{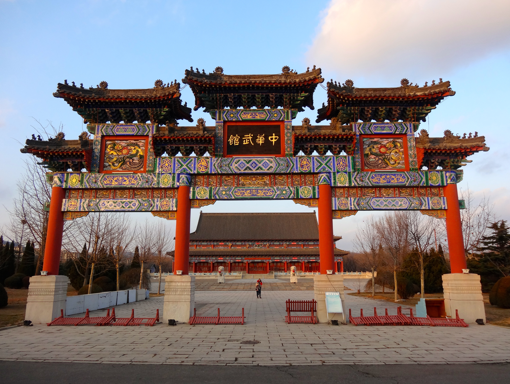}
        \end{subfigure}
        \\
        \begin{subfigure}[H]{\linewidth}
            \centering
    		\includegraphics[width=\linewidth]{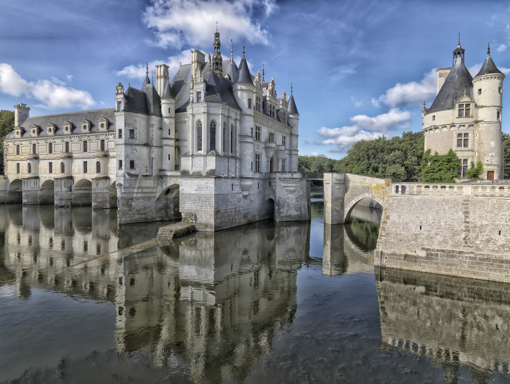}
        \end{subfigure}
        \subcaption*{Input}
    \end{subfigure}
    \begin{subfigure}[H]{0.15\linewidth}
        \centering
        \begin{subfigure}[H]{\linewidth}
            \centering
    		\includegraphics[width=\linewidth]{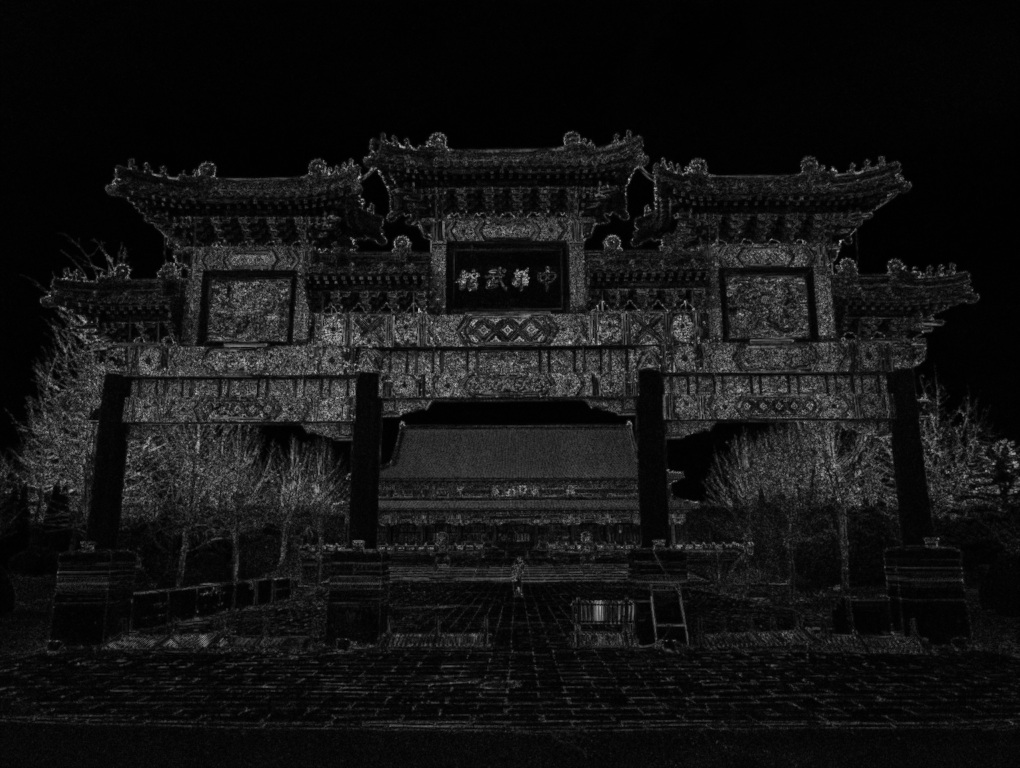}
        \end{subfigure}
        \\
        \begin{subfigure}[H]{\linewidth}
            \centering
    		\includegraphics[width=\linewidth]{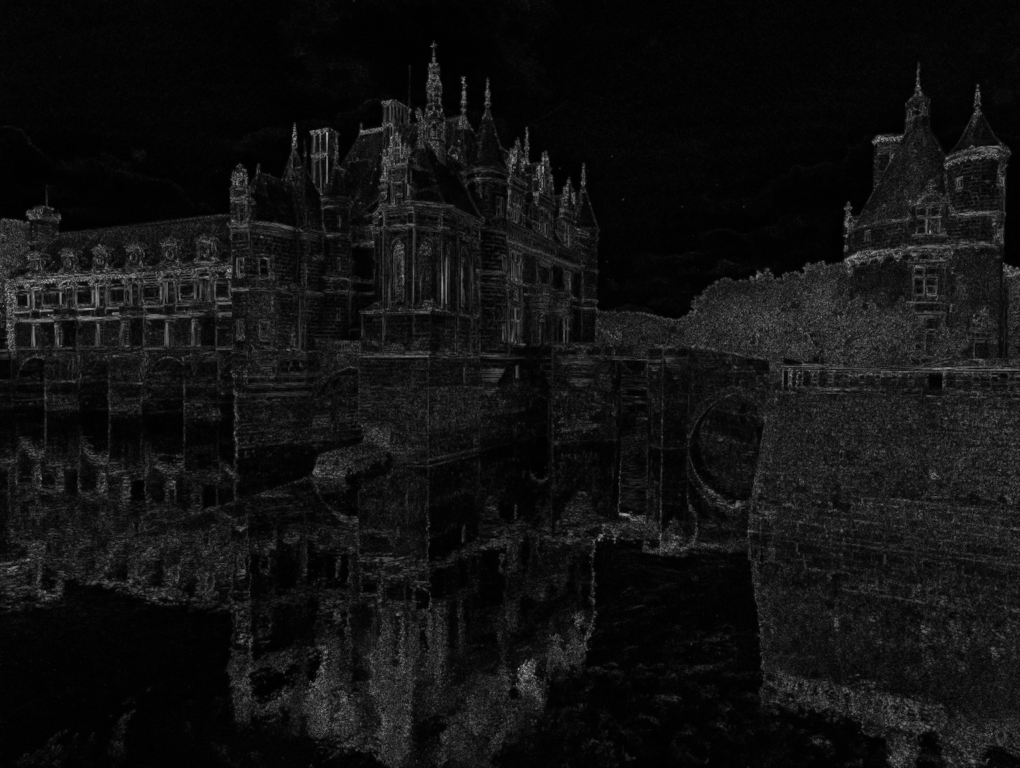}
        \end{subfigure}
        \subcaption*{1st}
    \end{subfigure}
    \begin{subfigure}[H]{0.15\linewidth}
        \centering
        \begin{subfigure}[H]{\linewidth}
            \centering
    		\includegraphics[width=\linewidth]{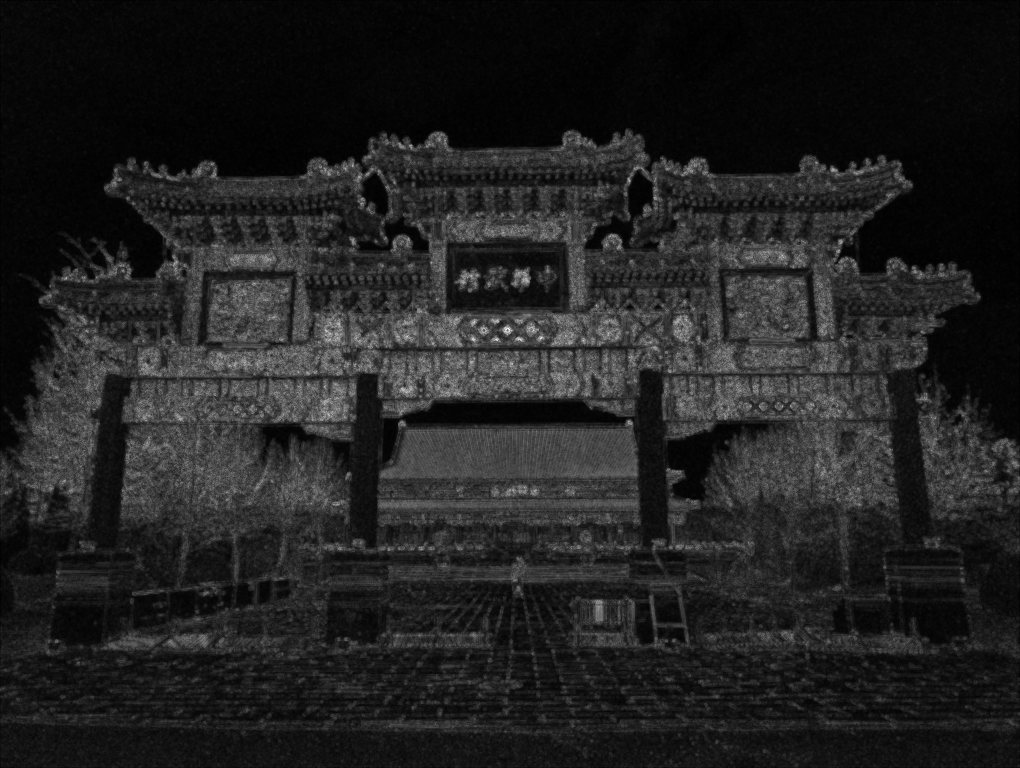}
        \end{subfigure}
        \\
        \begin{subfigure}[H]{\linewidth}
            \centering
    		\includegraphics[width=\linewidth]{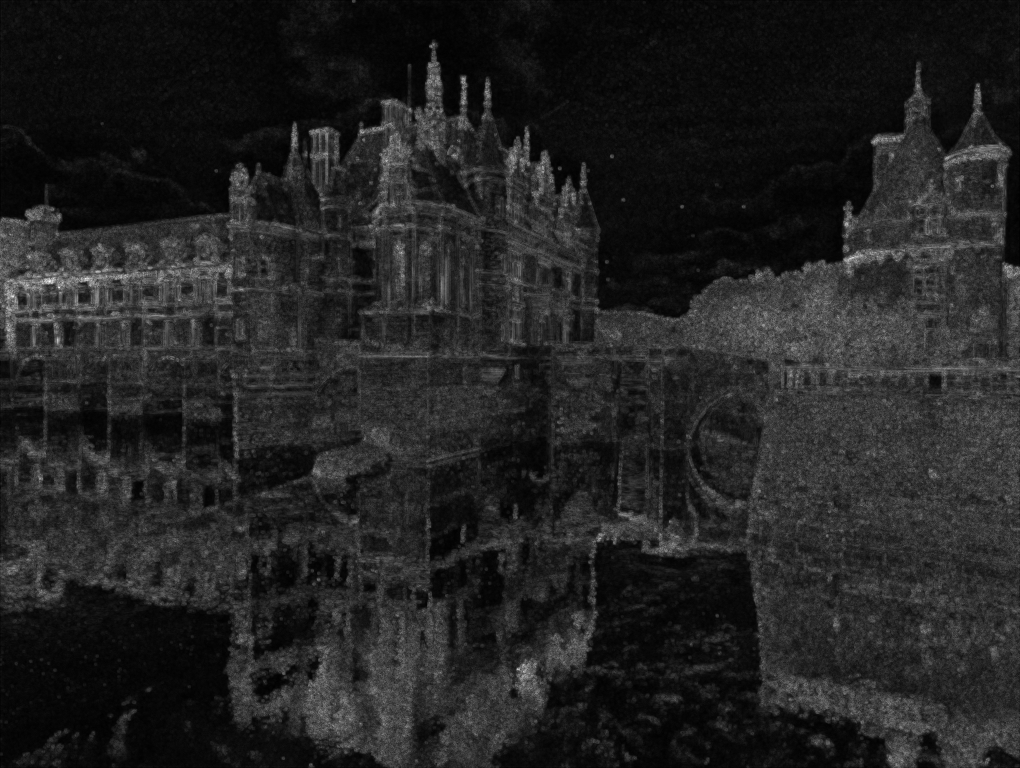}
        \end{subfigure}
        \subcaption*{3rd}
    \end{subfigure}
    \begin{subfigure}[H]{0.15\linewidth}
        \centering
        \begin{subfigure}[H]{\linewidth}
            \centering
    		\includegraphics[width=\linewidth]{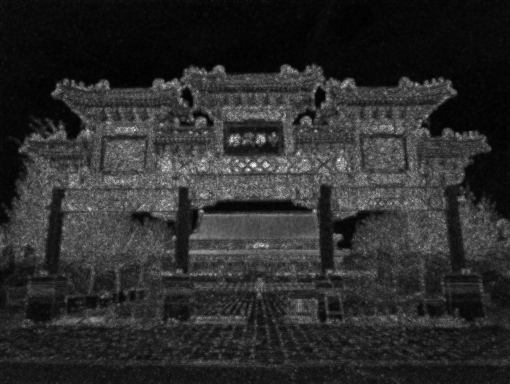}
        \end{subfigure}
        \\
        \begin{subfigure}[H]{\linewidth}
            \centering
    		\includegraphics[width=\linewidth]{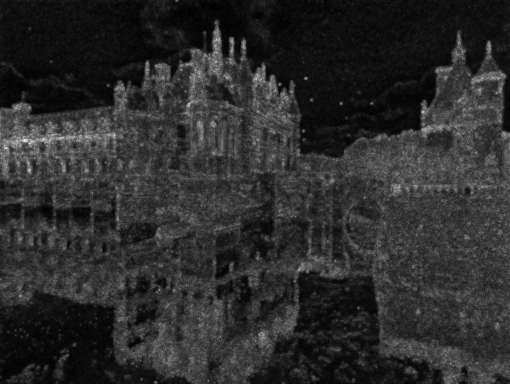}
        \end{subfigure}
        \subcaption*{5th}
    \end{subfigure}
    \begin{subfigure}[H]{0.15\linewidth}
        \centering
        \begin{subfigure}[H]{\linewidth}
            \centering
    		\includegraphics[width=\linewidth]{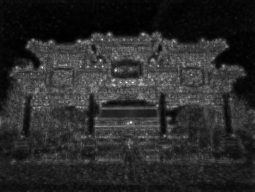}
        \end{subfigure}
        \\
        \begin{subfigure}[H]{\linewidth}
            \centering
    		\includegraphics[width=\linewidth]{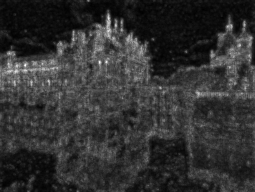}
        \end{subfigure}
        \subcaption*{9th}
    \end{subfigure}
    \begin{subfigure}[H]{0.15\linewidth}
        \centering
        \begin{subfigure}[H]{\linewidth}
            \centering
    		\includegraphics[width=\linewidth]{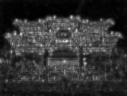}
        \end{subfigure}
        \\
        \begin{subfigure}[H]{\linewidth}
            \centering
    		\includegraphics[width=\linewidth]{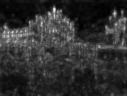}
        \end{subfigure}
        \subcaption*{13th}
    \end{subfigure}
    
    \caption{Visualization of extracted features from the 1st, 3rd, 5th, 9th and 13th layer of the pre-trained VGG-19.}
    \label{fig:vgg_features}
\vspace{-1.5em}
\end{figure*}

\subsection{Network Architecture} \label{sec:methods_architecture}
% In this section, we describe the Residual Local Feature Network (RLFN) in detail. 
The overall network architecture of our proposed Residual Local Feature Network (RLFN) is decipted in Figure \ref{fig:modelarchitecture}. Our RLFN mainly consists of three parts: the first feature extraction convolution, multiple stacked residual local feature blocks (RLFBs), and the reconstruction module. We denote $I_{LR}$ and $I_{SR}$ as the input and output of RLFN. In the first stage, we use a single $3\times3$ convolution layer to extract the coarse features:
\begin{equation}
  \begin{aligned}
   F_{0} = h_{ext}(I_{LR}),
  \end{aligned}
  \label{eqn:first_conv}
\end{equation}
where $h_{ext}(\cdot)$ denotes the convolution operation for feature extraction and $F_{0}$ is the extracted feature maps.
Then we use multiple RLFBs in a cascade manner for deep feature extraction. This process can be expressed by
\begin{equation}
  \begin{aligned}
  F_{n} = h_{RLFB}^{n}(h_{RLFB}^{n-1}(\dots h_{RLFB}^{0}(F_{0})\dots)),
  \end{aligned}
  \label{eqn:RLFBs}
\end{equation}
where $h_{RLFB}^{n}(\cdot)$ denotes the $n$-th RLFB function, and $F_{n}$ is the $n$-th output feature maps.

In addition, we use one $3\times3$ convolution layer to smooth the gradually refined deep feature maps. Next, the reconstruction module is applied to generate the final output $I_{SR}$.
\begin{equation}
  \begin{aligned}
  I_{SR} = f_{rec}((f_{smooth}(F_{n}) + F_{0})),
  \end{aligned}
  \label{eqn:RLFBs}
\end{equation}
where $f_{rec}(\cdot)$ represents the reconstruction module which consists of one single $3\times3$ convolution layer and one non-parametric sub-pixel operation. Besides, $f_{smooth}$ denotes a $3\times3$ convolution operation.

\textbf{Rethinking the RFDB} 
% \subsubsection{Rethinking the RFDB}
In this subsection, we rethink the efficiency of the residual feature distillation block (RFDB) proposed by RFDN.
As shown in Figure \ref{fig:RFDBarchitecture}, RFDB adopts a progressive feature refinement and distillation strategy in the beginning, and then use a $1\times1$ convolution for channel reduction. 
In the end, it applies an enhanced spatial attention (ESA) \cite{RFANet} layer and a residual connection.

Specifically, the feature refinement and distillation pipeline contains several steps. For each stage, RFDB adopts one refinement module that consists of one shallow residual block \cite{RFANet} (SRB) to refine the extracted features, and use a distillation module (a single $1\times1$ convolution layer) to distill features. Here we denote the refinement and distillation modules as $RM$ and $DM$, respectively. Given the input features $F_{in}$, the whole structure can be described by as
\begin{equation}
  \begin{aligned}
    F_{d_1}, F_{r_1} & = DM_{1}(F_{in}), RM_{1}(F_{in}) \\
    F_{d_2}, F_{r_2} & = DM_{2}(F_{r_1}), RM_{2}(F_{r_1}) \\
    F_{d_3}, F_{r_3} & = DM_{3}(F_{r_2}), RM_{2}(F_{r_2}) \\
    F_{d_4} & = DM_4(F_{r_3}),\\
  \end{aligned}
  \label{eqn:RFDB_1}
  \vspace{-0.2em}
\end{equation}
where $DM_{j}$, $RM_{j}$ denote the $j$-th distillation and refinement modules, respectively. $F_{d_j}$ represents the $j$-th distilled features, and $F_{r_j}$ is the $j$-th refined features that will be further processed by succeeding layers. Lastly, all the distilled features produced by previous distillation steps are concatenated together:
\begin{equation}
    F_{d} = Concat(F_{d_1}, F_{d_2}, F_{d_3}, F_{d_4}),
  \label{eqn:RFDB_1}
  \vspace{-0.2em}
\end{equation}
where $Concat$ represents the concatenation operation along the channel dimension.

Overall, RFDB utilizes progressive feature refinement together with multiple feature distillation connections for discriminative feature representations. Practically, the feature distillation connections implemented by several $1\times1$ convolution operations as well as a concatenation operation could significantly reduce the number of parameters as well as boost the restoration performance. However, this design severely deteriorates the inference speed.

We carefully analyze the efficiency of RFDB in Table \ref{tab:ab_block}. In particular, we remove the hierarchical distillation connections and create two variants of RFDB, namely, RFDB\_R\_48 (\textbf{R}efinement with convolution channel \textbf{48}) and RFDB\_R\_52 as shown in Figure \ref{fig:block_ablation}.  From Table \ref{tab:ab_block}, it is observed that RFDB\_R\_48 could reduce 25\% inference time compared to the original RFDB. Fortunately, the induced performance drop could be compensated by increasing the channel number of convolution layers. RFDB\_R\_52 surpasses RFDB\_R\_48 by a large margin in PSNR with just a slight increase regarding the inference time. More importantly, RFDB\_R\_52 yields comparable results with RFDB but shows great superiority in terms of inference speed. Therefore, in this work, we directly get rid of the feature distillation branch and make better use of the remaining progressive refinement on local features. 

\textbf{Residual Local Feature Block} 
% \subsubsection{Residual Local Feature Block} 
% Based on the consideration of reducing inference time and memory, RLFB (see \cref{fig:RLFBarchitecture}) extracts features by a sequential structure and minimize the number of inter-layer connections. Specifically, RLFB replaces feature aggregation in RFDB with an addition operation for local feature learning. 
% % To help understand the transition from RFDB to RLFB, we also draw an intermediate transition module in \cref{fig:Transitionarchitecture}. 
% We first remove the concatenation and the related feature distillation layers in RFDB to get the transition module. Then we move the position of addition in front of the $1\times1$ convolution, and we simplify SRB as a 3 $\times$ 3 convolution layer followed by a ReLU layer in order to reduce the operation as much as possible. 
In this subsection, we introduce the residual local feature block (RLFB) that could significantly reduce the inference time while the model capacity is maintained. As shown in Figure \ref{fig:RLFBarchitecture}, our proposed RLFB discards the multiple feature distillation connections, and only uses a few stacked CONV+RELU layers for local feature extraction. In particular, each feature refinement module in RLFB contains one $3\times3$ convolution layer followed by a ReLU activation function layer. Given the input features $F_{in}$, the whole structure is described by as

\begin{equation}
  \begin{aligned}
    F_{refined_1} & = RM_{1}(F_{in}), \\
    F_{refined_2} & = RM_{2}(F_{refined_1}) \\
    F_{refined_3} & = RM_{3}(F_{refined_2}),\\
    % F_{refined} & = F_{in} + F_{refined_3},\\
  \end{aligned}
  \label{eqn:RFDB_1}
  \vspace{-0.2em}
\end{equation}
where $RM_{j}$ denotes the $j$-th refinement module, and $F_{refined_j}$ is the $j$-th refined features. After multiple local feature refinement steps, we add the lastly refined features $F_{refined_3}$ with the skipped features $F_{in}$. Then we have
\begin{equation}
  \begin{aligned}
    F_{refined} & = F_{in} + F_{refined_3},\\
  \end{aligned}
  \label{eqn:RFDB_1}
  \vspace{-0.2em}
\end{equation}
where $F_{refined}$ is the final refined output features.

\begin{figure}[t]  
\vspace{-0.5em}
	\centering
	\includegraphics[width=0.9\linewidth,scale=1.0]{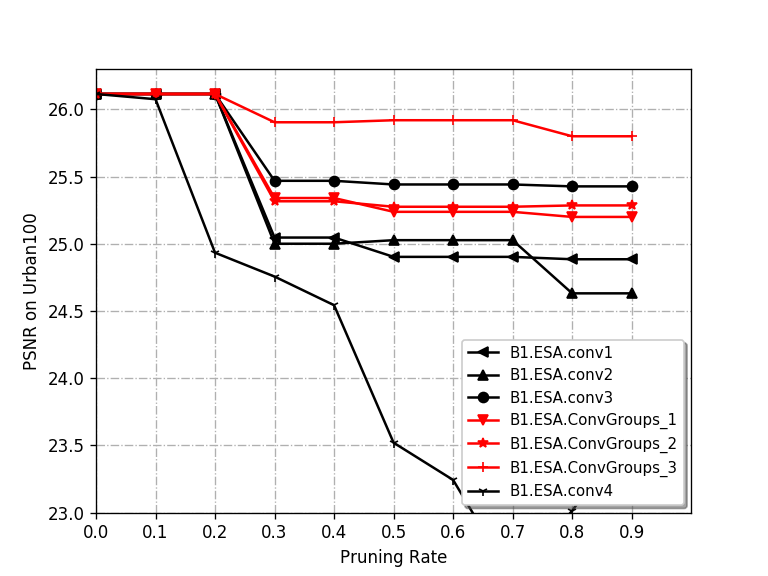}
    \caption{Parameter redundancy analysis of ESA modules on the Urban100 dataset. The red lines represent convolutional layers located in ConvGroups.}
	\label{fig:rfdb1_esa}
\vspace{-1.5em}
\end{figure}

Next, we follow RFDB to feed $F_{refined}$ to a $1\times1$ convolution layer and a subsequent ESA block to obtain the final output of RLFB. To further reduce the inference time, we developed a pruning sensitivity analysis tool based on one-shot structured pruning algorithm \cite{FP} to analyze the redundancy of ESA blocks in RFDB. As shown in Figure \ref{fig:rfdb1_esa}, the three convolution layers in ConvGroups rank top-1, top-3 and top-4 in redundancy, respectively.
Thus, for each ESA block, we reduce the number of convolution layers in ConvGroups to one. From Table \ref{tab:ab_esa}, we can find that this modification does not incur performance degradation. Instead, it brings a slight improvement regarding the inference time and model parameters.

% RFDB adopts ESA (see \cref{fig:ESAarchitecture}) to rescale the activations, so that the network can learn more discriminative features. ESA has a large receptive field, which ensures the accuracy of spatial attention modeling. Although our experimental results also verify that it has better performance than the attention modules commonly used in high-level tasks\cite{SE,CBAM,SRM,TRIP}, we believe that it contains too many convolutional layers and still has a compressible space. Pruning sensitivity can represent parameter redundancy\cite{LWC}, so we developed a pruning sensitivity analysis tool based on one-shot structured pruning algorithm\cite{FP} to analyze the redundancy of ESA modules in RFDB. As shown in \cref{fig:rfdb1_esa}, the three convolution layers in ConvGroups are in the top four in redundancy. Based on this finding, we replace ConvGroups with one convolution layer in RLFB, which has no effect on performance. See \ref{sec:arch_optim} for more details.

\subsection{Revisiting the Contrastive Loss}  \label{sec:methods_cl}
Contrastive learning has shown impressive performance in self-supervised learning\cite{CLSurvey, SimCLR, MoCo, SimCLRV2, MoCoV2}. The basic idea behind is to push positives closer to anchors, and push negatives away from anchors in the latent space. Recent works \cite{AECR-Net, CSD} propose a novel contrastive loss, and demonstrate its effectiveness by improving the quality of reconstructed images. The contrastive loss is defined as:
\begin{equation}
% \vspace{-1em}
    \label{eqn: contrastive_loss}
    CL = \sum_{i = 1}^{n}\lambda_{i}\frac{d(\phi_{i}(Y_{anchor}), \phi_{i}(Y_{pos}))}{d(\phi_{i}(Y_{anchor}), \phi_{i}(Y_{neg}))},
\vspace{-1em}
\end{equation}
where $\phi_{j}$ denotes the intermediate features from the $j$-th layer. $d(x, y)$ is the L1-distance between $x$ and $y$, and $\lambda_{j}$ is the balancing weight for each layer. AECR-Net\cite{AECR-Net} and CSD\cite{CSD} extract the features from the 1st, 3rd, 5th, 9th and 13th layers of the pre-trained VGG-19. However, we experimentally find that the PSNR is decreased when we employ the contrastive loss.

We next try to investigate its reason to explain the discrepancy. The contrastive loss defined in \cref{eqn: contrastive_loss} mainly depends on the difference of feature maps between two images $Y_{1}$ and $Y_{2}$. Therefore, we try to visualize the difference map of their feature maps extracted by a pre-trained model $\phi$:
\begin{equation}
% \vspace{1em}
    \label{eqn: difference_map}
    DMAP_{i, j} = \sqrt{\sum_{k=1}^{K}(\phi(Y_{1})_{i, j, k} - \phi(Y_{2})_{i, j, k})^{2}},
\end{equation}
% where $i, j, k$ are the spatial coordinates of $Y_{1}$ and $Y_{2}$. 
where $i, j$ are the spatial coordinates of $Y_{1}$ and $Y_{2}$, while $k$ is the channel index of $Y_{1}$ and $Y_{2}$. We use the 100 validation high-resolution images in DIV2K dataset as $Y_{1}$, the corresponding images degraded blur kernels as $Y_{2}$.
Figure \ref{fig:vgg_features} presents visualization examples. A surprising observation is that the difference map of features extracted from deeper layers are more semantic, but lacking accurate details. For example, the edges and textures are mostly preserved by the 1st layer, while features from the 13th layer only preserve the overall spatial structure and details are generally missing. In summary, features from deep layers can improve the performance in terms of real perceptual quality because it provides more semantic guidance. Features from shallow layers preserve more accurate details and textures, which are critical for PSNR-oriented models. It suggests that we should utilize features from shallow layers to improve the PSNR of the trained model.

% Our work is partly motivated by previous work \cite{PerceptualLoss}. It present similar conclusion that images reconstructed from deeper layers has high-quality results on image style transfer. However, there is no comprehensive study of the root cause. 

\begin{figure}[!htb]
% \vspace{1em}
    \centering
    \begin{subfigure}[H]{0.45\linewidth}
        \centering
        \begin{subfigure}[H]{\linewidth}
            \centering
    		\includegraphics[width=\linewidth]{dm_0826_relu1_1_p.png}
        \end{subfigure}
        \\
        \begin{subfigure}[H]{\linewidth}
            \centering
    		\includegraphics[width=\linewidth]{dm_0865_relu1_1_p.png}
        \end{subfigure}
        \subcaption{$DMAP_{p}$}
    \end{subfigure}
    \begin{subfigure}[H]{0.45\linewidth}
        \centering
        \begin{subfigure}[H]{\linewidth}
            \centering
    		\includegraphics[width=\linewidth]{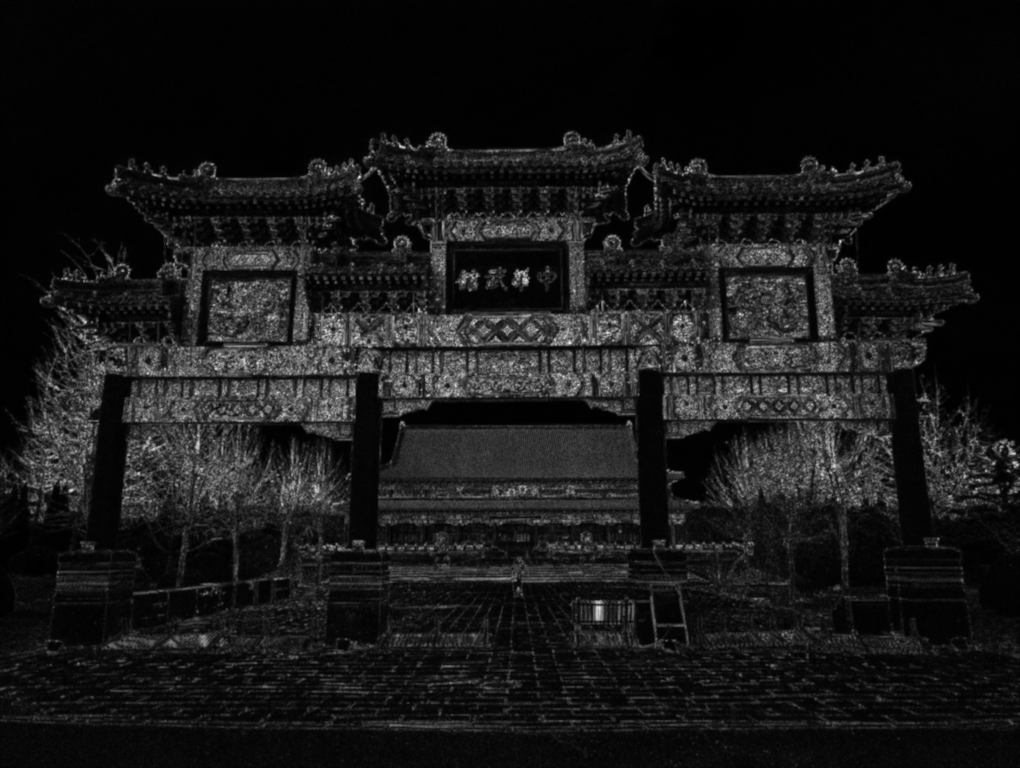}
        \end{subfigure}
        \\
        \begin{subfigure}[H]{\linewidth}
            \centering
    		\includegraphics[width=\linewidth]{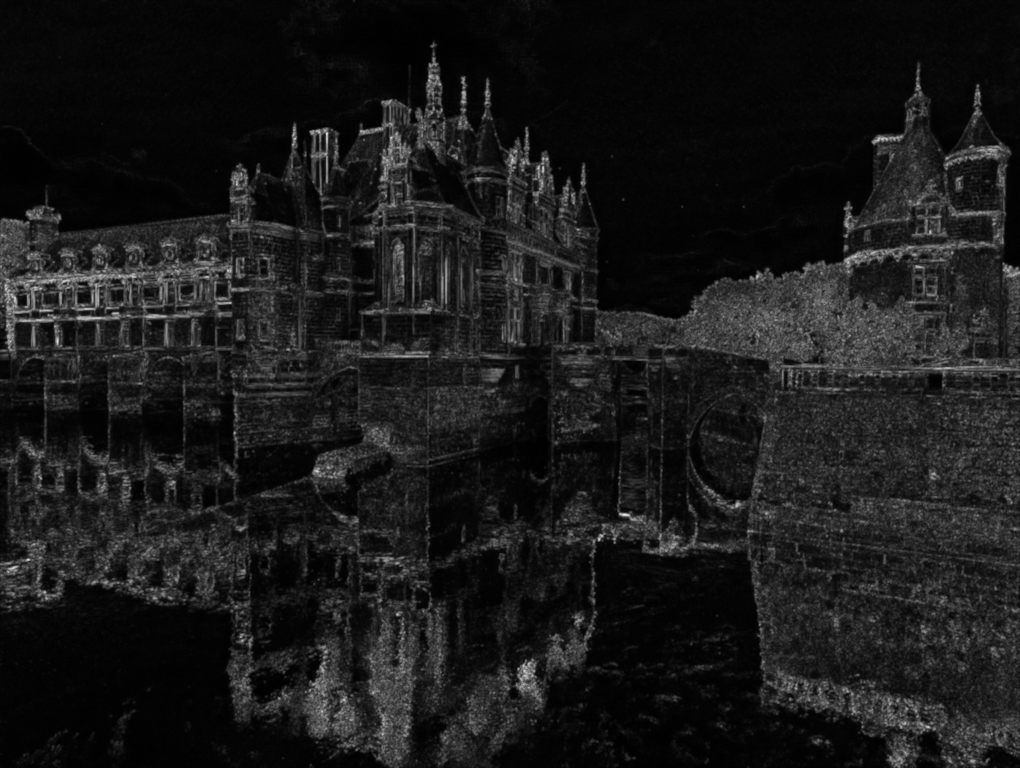}
        \end{subfigure}
        \subcaption{$DMAP_{r}$}
    \end{subfigure}
    \caption{The difference maps of the pre-trained VGG-19 $DMAP_{p}$ and our proposed feature extractor $DMAP_{r}$. $DMAP_{r}$ has stronger response and can capture more details and textures compared with $DMAP_{p}$.}
    \label{fig:vgg_tanh}
\vspace{-1em}
\end{figure}

To further improve the contrastive loss, we revisit the architectures of the feature extractor. The original contrastive loss tries to minimize the distance between two activated features after the ReLU activation function. However, the ReLU function is unbounded above and the activated feature map is sparse, which results in loss of information and provides weaker supervision. Therefore, we replace the ReLU activation function of the feature extractor with the Tanh function. 

\begin{table*}[!htb]
% \vspace{1em}
    % \renewcommand{\arraystretch}{1.2}
    \centering
    % \scalebox{0.65}{
    \resizebox{0.9\linewidth}{!}{
        \begin{tabular}{c c c c c c c c}
            \toprule
            \multirow{2}{*}{Scale} & \multirow{2}{*}{Model} & Params & Runtime & Set5 & Set14 & BSD100 & Urban100 \\
            ~ & ~ & (K) & (ms) &PSNR$\uparrow$ / SSIM$\uparrow$ & PSNR$\uparrow$ / SSIM$\uparrow$ & PSNR$\uparrow$ / SSIM$\uparrow$ & PSNR$\uparrow$ / SSIM$\uparrow$\\
     
            \midrule
            \multirow{11}{*}{$\times$ 2} & SRCNN\cite{SRCNN} & 24 & 6.92 
            & 36.66 / 0.9542
            & 32.42 / 0.9063
            & 31.36 / 0.8879
            & 29.50 / 0.8946 \\
            % & 27.58 / 0.8555 \\
            
            ~ & FSRCNN\cite{FSRCNN} & 12 & 9.02 
            & 36.98 / 0.9556
            & 32.62 / 0.9087
            & 31.50 / 0.8904
            & 29.85 / 0.9009 \\
            % & 27.90 / 0.8610 \\
            
            ~ & VDSR\cite{VDSR} & 666 & 35.37 
            & 37.53 / 0.9587
            & 33.05 / 0.9127
            & 31.90 / 0.8960
            & 30.77 / 0.9141 \\
            % & 28.83 / 0.8870 \\
            
            ~ & DRCN\cite{DRCN} & 1774 & 716.45 
            & 37.63 / 0.9588 
            & 33.04 / 0.9118 
            & 31.85 / 0.8942 
            & 30.75 / 0.9133\\
            % & 28.93 / 0.8854 \\
            
            ~ & LapSRN\cite{LapSRN} & 251 & 53.98 
            & 37.52 / 0.9591 
            & 32.99 / 0.9124 
            & 31.80 / 0.8952 
            & 30.41 / 0.9103 \\
            % & 29.09 / 0.8900 \\
            
            ~ & CARN\cite{CARN} & 1592 & 159.10 
            & 37.76 / 0.9590 
            & 33.52 / 0.9166 
            & 32.09 / 0.8978 
            & 31.92 / 0.9256 \\
            % & 30.47 / 0.9084 \\
            
            ~ & IMDN\cite{IMDN} & 694 & 77.34 
            & 38.00 / 0.9605 
            & 33.63 / 0.9177 
            & \textcolor{blue}{32.19} / 0.8996 
            &  \textcolor{blue}{32.17} / 0.9283\\
            % & 30.45 / 0.9075 \\
            
            ~ & RFDN\cite{RFDN} & 534 & 74.51 
            & \textcolor{blue}{38.05} / 0.9606 
            & \textcolor{blue}{33.68} / \textcolor{blue}{0.9184} 
            & 32.16 / 0.8994 
            & 32.12 / 0.9278 \\
            % & 30.58 / 0.9089 \\
            
            % ~ & RFDN-L\cite{RFDN} & 626
            % & 38.08 / 0.9606 
            % & 33.67 / 0.9190 
            % & 32.18 / 0.8996 
            % & 32.24 / 0.9290 \\
            % % & 30.61 / 0.9096 \\
            
            ~ & MAFFSRN\cite{MAFFSRN} & 402 & 152.91 
            & 37.97 / 0.9603 
            & 33.49 / 0.9170 
            & 32.14 / 0.8994 
            & 31.96 / 0.9268 \\
            
            ~ & ECBSR\cite{ECBSR} & 596 & 39.96 
            & 37.90 / \textcolor{red}{0.9615}
            & 33.34 / 0.9178
            & 32.10 / \textcolor{red}{0.9018}
            & 31.71 / 0.9250 \\
            
            ~ & FDIWN-M\cite{FDIWN} & - & - 
            & - / -
            & - / -
            & - / -
            & - / - \\

            ~ & \textbf{RLFN-S (ours)} & 454 & 56.09 
            & \textcolor{blue}{38.05} / \textcolor{blue}{0.9607}
            & \textcolor{blue}{33.68} / 0.9172
            & \textcolor{blue}{32.19} / 0.8997
            & \textcolor{blue}{32.17} / \textcolor{blue}{0.9286} \\
            % & 31.70 / \textcolor{blue}{0.9339} \\
            
            ~ & \textbf{RLFN (ours)} & 527 & 60.39  
            & \textcolor{red}{38.07} / \textcolor{blue}{0.9607}
            & \textcolor{red}{33.72} / \textcolor{red}{0.9187}
            & \textcolor{red}{32.22} / \textcolor{blue}{0.9000}
            & \textcolor{red}{32.33} / \textcolor{red}{0.9299} \\
            % & 31.81 / \textcolor{red}{0.9354} \\
        
            % \midrule
            
            \midrule
            \multirow{11}{*}{$\times$ 4} & SRCNN\cite{SRCNN} & 57 & 1.90
            & 30.48 / 0.8628
            & 27.49 / 0.7503
            & 26.90 / 0.7101
            & 24.52 / 0.7221 \\
            % & 27.58 / 0.8555 \\
            
            ~ & FSRCNN\cite{FSRCNN} & 13 & 2.22 
            & 30.72 / 0.8660
            & 27.61 / 0.7550
            & 26.98 / 0.7150
            & 24.62 / 0.7280 \\
            % & 27.90 / 0.8610 \\
            
            ~ & VDSR\cite{VDSR} & 666 & 8.95 
            & 31.35 / 0.8838 
            & 28.01 / 0.7674
            & 27.29 / 0.7251
            & 25.18 / 0.7524 \\
            % & 28.83 / 0.8870 \\
            
            ~ & DRCN\cite{DRCN} & 1774 & 176.59 
            & 31.53 / 0.8854
            & 28.02 / 0.7670 
            & 27.23 / 0.7233
            & 25.14 / 0.7510 \\
            % & 28.93 / 0.8854 \\
            
            ~ & LapSRN\cite{LapSRN} & 502 & 66.81 
            & 31.54 / 0.8852 
            & 28.09 / 0.7700
            & 27.32 / 0.7275 
            & 25.21 / 0.7562 \\
            % & 29.09 / 0.8900 \\
            
            ~ & CARN\cite{CARN} & 1592 & 39.96 
            & 32.13 / 0.8937 
            & 28.60 / 0.7806 
            & 27.58 / 0.7349 
            & 26.07 / 0.7837 \\
            % & 30.47 / 0.9084 \\
            
            ~ & IMDN\cite{IMDN} & 715 & 20.56 
            & 32.21 / 0.8948 
            & 28.58 / 0.7811 
            & 27.56 / 0.7353 
            & 26.04 / 0.7838 \\
            % & 30.45 / 0.9075 \\
            
            ~ & RFDN\cite{RFDN} & 550 & 20.40 
            & \textcolor{red}{32.24} / \textcolor{blue}{0.8952}
            & \textcolor{blue}{28.61} / \textcolor{red}{0.7819}
            & 27.57 / 0.7360
            & 26.11 / 0.7858 \\
            % & 30.58 / 0.9089 \\
            
            % ~ & RFDN-L\cite{RFDN} & 643K
            % & 32.28 / 0.8957
            % & 28.61 / 0.7818 
            % & 27.58 / 0.7363
            % & 26.20 / 0.7883 \\
            % % & 30.61 / 0.9096 \\
            
            ~ & MAFFSRN\cite{MAFFSRN} & 441 & 39.69 
            & 32.18 / 0.8948 
            & 28.58 / 0.7812 
            & 27.57 / \textcolor{blue}{0.7361} 
            & 26.04 / 0.7848 \\
            
            ~ & ECBSR\cite{ECBSR} & 603 & 10.21 
            & 31.92 / 0.8946 
            & 28.34 / 0.7817 
            & 27.48 / 0.7393
            & 25.81  /0.7773 \\
            
            ~ & FDIWN-M\cite{FDIWN} & 454 & - 
            & 32.17 / 0.8941 
            & 28.55 / 0.7806 
            & \textcolor{blue}{27.58} / \textcolor{red}{0.7364}
            & 26.02 / 0.7844 \\
            
            % ~ & FDIWN\cite{FDIWN} & 664
            % & 32.23 / 0.8955 
            % & 28.66 / 0.7829 
            % & 27.62 / 0.7380 
            % & 26.28 / 0.7919 \\
            
            ~ & \textbf{RLFN-S (ours)} & 470 & 15.16 
            & \textcolor{blue}{32.23} / \textcolor{red}{0.8961}
            & \textcolor{blue}{28.61} / \textcolor{blue}{0.7818}
            & \textcolor{blue}{27.58} / 0.7359
            & \textcolor{blue}{26.15} / \textcolor{blue}{0.7866} \\
            
            ~ & \textbf{RLFN (ours)} & 543 & 16.41 
            & \textcolor{red}{32.24} / \textcolor{blue}{0.8952}
            & \textcolor{red}{28.62} / 0.7813
            & \textcolor{red}{27.60} / \textcolor{red}{0.7364}
            & \textcolor{red}{26.17} / \textcolor{red}{0.7877} \\
            \bottomrule
            \end{tabular}
    % }
    }
    \caption{Quantitative results of the state-of-the-art models on four benchmark datasets. The best and second-best results are marked in \textcolor{red}{red} and \textcolor{blue}{blue} colors, respectively.}
    \label{tab:quantitative_sota}
\vspace{-1.5em}
\end{table*}

Moreover, since the VGG-19 is trained with the ReLU activation function, the performance is not guaranteed if the ReLU activation is replaced with the Tanh function without any training. Some recent works\cite{GenericPerceptualLoss, RandomStyleTransfer} shows that a randomly initialized network with good architecture is sufficient to capture perceptual details. Inspired by these works, we build a randomly initialized two-layer feature extractor, which has an architecture of Conv\_k3s1-Tanh-Conv\_k3s1. The difference maps of the pre-trained VGG-19 and our proposed feature extractor are presented in Figure \ref{fig:vgg_tanh}. We can observe that the difference map of our proposed feature extractor has stronger response and can capture more details and textures, compared with the difference map of the pre-trained VGG-19. This also provides evidence that a randomly initialized feature extractor can already capture some structural information and pre-training is not necessary.

\subsection{Warm-Start Strategy}  \label{sec:methods_ws}
For large scale factors like 3 or 4 in the SR task, some previous works\cite{RFDN} use the 2x model as a pre-trained network instead of training them from scratch. The 2x model provides good initialized weights which accelerates the convergence and improves the final performance. However, we can only enjoy the benefit once because the scale factors of pre-trained models and target models are different.

% To solve the limitation, 
To address this issue,
we propose a novel multi-stage warm-start training strategy, which can empirically improve the performance of SISR models. In the first stage, we train RLFN from scratch. Then in the next stage, instead of training from scratch, we load the weights of RLFN of the previous stage, which is referred to as the warm-start policy. The training settings, such as batch size and the learning rate, follow exactly the same training scheme in the first stage. In the following of this paper, we use \textbf{RFLN\_ws\_$i$} to denote the trained model which employs warm-start $i$ times (after $i+1$ stages). For example, RFLN\_ws\_$1$ denotes a two-stage training process. In the first stage, we train RLFN from scratch. Then in the second stage, RLFN loads the pre-trained weights and is trained following the same training scheme as the first stage.

\section{Experiments}
\subsection{Setup}
% \subsubsection{Datasets and Metrics}
\noindent\textbf{Datasets and Metrics} 
We use the 800 training images in DIV2K dataset\cite{DIV2K} for training. We test the performance of our models on four benchmark dataset: Set5\cite{Set5}, Set14\cite{Set14}, BSD100\cite{BSD100} and Urban100\cite{Urban100}. We evaluate the PSNR and SSIM on the Y channel of YCbCr space.

\begin{figure*}[!htb]
  \centering
  \subfloat[][RFDB]{
    \includegraphics[width=0.24\linewidth,height=14em, ]{a_RFDB}
    \label{fig:RFDB_lab}
  }
%   \qquad
  \subfloat[][RFDB\_R\_48]{
    \includegraphics[width=0.16\linewidth,height=14em,
    ]{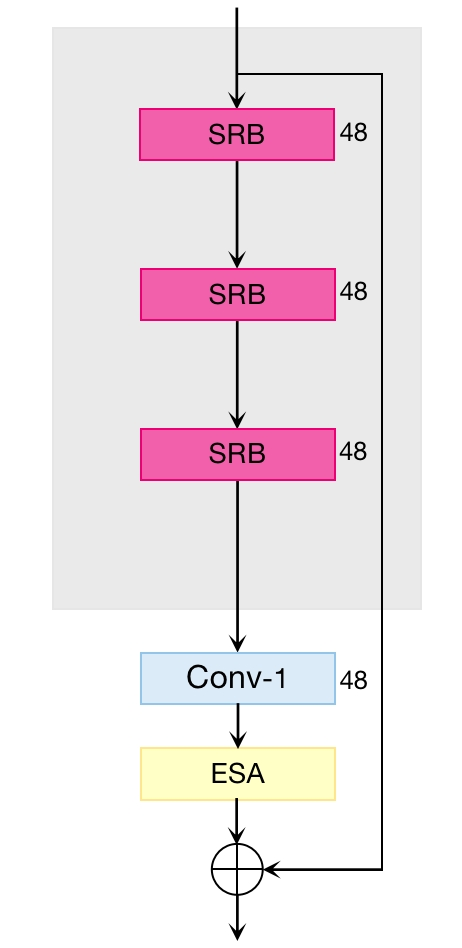}
    \label{fig:Transitionar1}
  }
%   \qquad
  \subfloat[][RFDB\_R\_52]{
    \includegraphics[width=0.16\linewidth,height=14em,
    ]{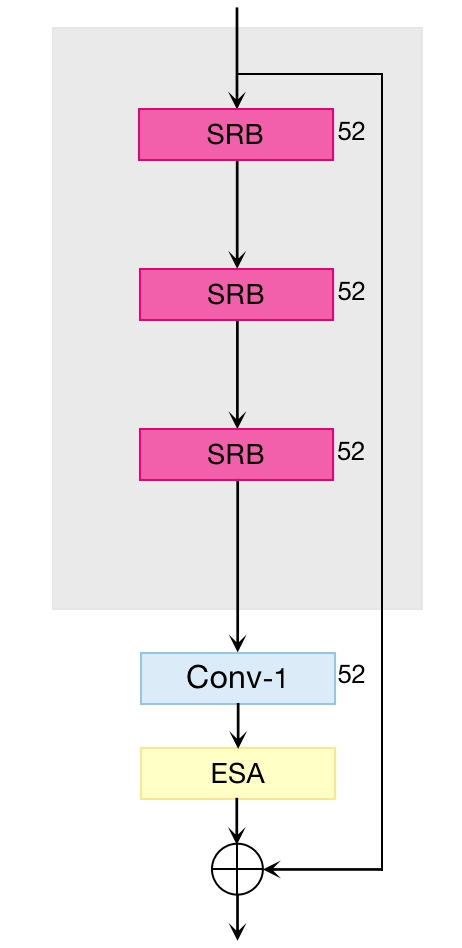}
    \label{fig:Transition2}
  }
%   \qquad
  \subfloat[][RLFB]{
    \includegraphics[width=0.16\linewidth,height=14em,
    ]{c_RLFB}
    \label{fig:RLFB_lab}
  }
  
%  \vspace{1em}
  \caption{The blocks used in ablation study. RFDB\_R represent the refinement part of RFDB.}
\label{fig:block_ablation}
\vspace{-1.5em}
\end{figure*}

% \subsubsection{Training Details}
\noindent\textbf{Training Details}
Our models are trained on RGB channels and we augment the training data with random flipping and 90 degree rotations. LR images are generated by down-sampling HR images with bicubic interpolation in MATLAB. We randomly crop HR patches of size $256\times256$ from ground truth, and the mini-batch size is set to 64. The training process has three stages. In the first stage, we train the model from scratch. Then we employ the warm-start strategy twice. 
In each stage, we adopt Adam optimizer\cite{Adam} by setting $\beta_{1}=0.9$, $\beta_{2}=0.999$ and $\epsilon=10^{-8}$ and minimize the L1 loss following the training process of RFDN\cite{RFDN}. The initial learning rate is 5e-4, and is halved every $2\times10^{5}$ iterations. Moreover, we additionally employ the widely used contrastive loss\cite{CSD} in the third stage. %when training RLFN\_ws\_2. 
This training strategy is adopted in the following of this paper unless otherwise stated.
We implement two models, RLFN-S and RLFN. The number of RLFB is set to 6 in both models. We set the number of channels of RLFN to 52. To achieve better runtime, RLFN-S has a smaller channel number of 48.

\subsection{Quantitative Results}
We compare our models with several advanced efficient super-resolution models\cite{FSRCNN, VDSR, LapSRN, IMDN, RFDN, MAFFSRN, ECBSR, FDIWN} with scale factor of 2 and 4. The quantitative performance comparison on several benchmark datasets is shown in Table~\ref{tab:quantitative_sota}. Compared with other state-of-the-art models, the proposed RLFN-S and RLFN achieve superior performance in terms of both PSNR and SSIM. RLFN-S can achieve comparable or even better performance than RFDN\cite{RFDN}, with 80K less parameters. With similar model size, RLFN outperforms other methods by a large margin on all benchmark datasets.
We also visualize the trade-off between performance and inference time, and the trade-off between performance and parameters in Figure \ref{fig:comparison}, respectively.
The inference time in Figure \ref{fig:comparison}  is the average of 10 runs with CUDA Toolkit 9.0.176 on the NVIDIA 1080Ti GPU.  From Figure \ref{fig:comparison} we can see that our RLFN obtains a better trade-off between quality and inference time than other existing methods.

\subsection{Ablation Study}

\noindent\textbf{Effectiveness of Architecture Optimization}
% \subsubsection{Effectiveness of Architecture Optimization}
\label{sec:arch_optim}
To evaluate the effectiveness of our model architecture optimization, we design two variants of RFDB. As shown in Figure \ref{fig:block_ablation}, we remove the feature distillation layers in RFDB to get RFDB\_R\_48, then RFDB\_R\_52 increases the number of channel to 52 and the middle channel in ESA to 16 for reducing the performance drop,  and RLFB removes the intensive addition operations inside SRB based on RFDB\_R\_52. RFDB, RFDB\_R\_48, RFDB\_R\_52, and RLFB are stacked as the body part of the SR network (Figure \ref{fig:modelarchitecture}) and trained with the settings of the first stage described in Section 4.1. As shown in Table~\ref{tab:ab_block}, RLFB maintains the same level of restoration performance with RFDB but has obvious speed advantages. %which demonstrates the efficiency of our model structure.

We also investigate the effect of reducing convolution layers
in ESA ConvGroups in the same setting. RLFB\_esa\_g3 and RLFB maintains the similar  restoration performance in Table~\ref{tab:ab_esa}, which means reducing two layers in ESA ConvGroups does not sacrifice performance, but accelerates the network inference. 

\begin{table}[!htb]
% \vspace{1em}
% \renewcommand{\arraystretch}{1.2}
    \centering
    \scalebox{0.55}{
        \begin{tabular}{c c c c c c c }
            \toprule
            \multirow{2}{*}{Model} & Params & {Runtime} & Set5 & Set14 & BSD100 & Urban100 \\
            ~ & (K) & (ms) & PSNR / SSIM & PSNR / SSIM & PSNR / SSIM & PSNR / SSIM\\
            
            \midrule
            
            % RFDN & 550 & 75.7
            % & 32.24 / 0.8953
            % & 28.59 / 0.7814
            % & 27.54 / 0.7355
            % & 26.15 / 0.7868 \\

            RFDB & 568 & 82.3 
            & 32.22 / 0.8949
            & 28.64 / 0.7824
            & 27.58 / 0.7361
            & 26.09 / 0.7859 \\
            
            RFDB\_R\_48 & 470 & 61.7 
            & 32.12 / 0.8942
            & 28.59 / 0.7806
            & 27.54 / 0.7351
            & 26.00 / 0.7832 \\
            
            RFDB\_R\_52 & 572 & 67.0 
            & 32.20 / 0.8948
            & 28.62 / 0.7817
            & 27.57 / 0.7358
            & 26.11 / 0.7861 \\
            
            RLFB(ours) & 543 & 63.2
            & 32.22 / 0.8952
            & 28.61 / 0.7817
            & 27.58 / 0.7359
            & 26.11 / 0.7865 \\
            
            \bottomrule
        \end{tabular}
    }
    \caption{Comparison of RFDB, its two variants: RFDB\_R\_48, RFDB\_R\_52, and our RLFN for 4x SR. Runtime is the average of 10 runs on DIV2K validation set.}
    \label{tab:ab_block}
\vspace{-1.5em}
\end{table}

\begin{table}[!htb]
% \vspace{1em}
% \renewcommand{\arraystretch}{1.2}
    \centering
    \scalebox{0.55}{
        \begin{tabular}{c c c c c c c }
            \toprule
            \multirow{2}{*}{Model} & Params & {Runtime} & Set5 & Set14 & BSD100 & Urban100 \\
            ~ & (K) & (ms) & PSNR / SSIM & PSNR / SSIM & PSNR / SSIM & PSNR / SSIM\\
            
            \midrule
            
            RLFB\_esa\_g3 & 572 & 63.6
            & 32.20 / 0.8952 
            & 28.61 / 0.7818 
            & 27.57 / 0.7363 
            & 26.12 / 0.7869 \\
            
            RLFB (ours) & 543 & 63.2
            & 32.22 / 0.8952
            & 28.61 / 0.7817
            & 27.58 / 0.7359
            & 26.11 / 0.7865 \\
            
            % RLFB (ours) & 543 & 63.2
            % & 32.24 / 0.8953
            % & 28.59 / 0.7815
            % & 27.57 / 0.7360
            % & 26.12 / 0.7872 \\
            
            \bottomrule
        \end{tabular}
    }
    \caption{Comparison of RLFB\_esa\_g3 and our RLFB for 4$\times$ SR. RLFB\_esa\_g3 uses three convolution layers in ConvGroups, while our RLFB uses just one. Runtime is the average of 10 runs on DIV2K validation set for 4x SR.}
    \label{tab:ab_esa}
\vspace{-1em}
\end{table}

\noindent\textbf{Effectiveness of Contrastive Loss}
% \subsubsection{Effectiveness of Contrastive Loss}
To investigate the effectiveness of contrastive loss, we remove the contrastive loss in the second warm-start stage and only employs L1 loss. As shown in Table~\ref{tab:ab_cl}, the contrastive loss consistently improves the performance in terms of both PSNR and SSIM on four benchmark datasets.

\begin{table}[!htb]
% \vspace{1em}
% \renewcommand{\arraystretch}{1.2}
    \centering
    \scalebox{0.63}{
        \begin{tabular}{c c c c c}
            \toprule
            \multirow{2}{*}{Model} & Set5 & Set14 & BSD100 & Urban100 \\
            ~ & PSNR / SSIM & PSNR / SSIM & PSNR / SSIM & PSNR / SSIM\\
            
            \midrule
            RLFN-S\_ws\_2
            & 32.22 / 0.8960
            & 28.60 / 0.7818
            & 27.57 / 0.7359
            & 26.13 / 0.7865 \\

            RLFN-S\_ws\_2 + $CL$
            & 32.23 / 0.8961
            & 28.61 / 0.7818
            & 27.58 / 0.7359
            & 26.15 / 0.7866 \\
            \bottomrule
        \end{tabular}
    }
    \caption{Effect of contrastive loss for 4x SR. RLFN-S\_ws\_2 applies warm-start twice and RLFN-S\_ws\_2 + $CL$ employs contrastive loss in the second warm-start stage. Contrastive loss improves the performance in terms of both PSNR and SSIM.}
    \label{tab:ab_cl}
\vspace{-1em}
\end{table}

\noindent\textbf{Effectiveness of Warm-Start Strategy}
% \subsubsection{Effectiveness of Warm-Start Strategy}
To demonstrate the effectiveness of our proposed warm-start strategy, we compare RLFN-S\_ws\_1 as the baseline and two variants of different learning rate strategies, RLFN-S\_e2000 and RLFN-S\_clr. Contrastive loss is not used in this comparison while other training settings remain the same. They set the total epochs to 2000 to be compared with RLFN-S\_ws\_1. RLFN-S\_e2000 halves the learning rate every $4\times10^{5}$ iterations. RLFN-S\_clr applies a cyclical learning rate policy, which is the same as RLFN-S\_ws\_1. However, it loads the state of the optimizer while RLFN-S\_ws\_1 applies the default initialization. As shown in Table~\ref{tab:ab_training_strategy}, RLFN-S\_e2000 and RLFN-S\_clr decrease PSNR and SSIM compared with our proposed warm-start strategy. It indicates that the warm-start strategy helps to jump out of the local minimum during the optimization process and improve the overall performance. 

\begin{table}[htbp]
% \vspace{1em}
% \renewcommand{\arraystretch}{1.2}
    \centering
    \scalebox{0.65}{
        \begin{tabular}{c c c c c}
            \toprule
            \multirow{2}{*}{Model} & Set5 & Set14 & BSD100 & Urban100 \\
            ~ & PSNR / SSIM & PSNR / SSIM & PSNR / SSIM & PSNR / SSIM\\
            
            \midrule
            RLFN-S\_e2000
            & 32.17 / 0.8953
            & 28.58 / 0.7815
            & \textcolor{blue}{27.57} / 0.7354
            & 26.08 / 0.7849 \\
            
            RFLN-S\_clr
            & \textcolor{blue}{32.20} / \textcolor{blue}{0.8959}
            & \textcolor{blue}{28.59} / \textcolor{blue}{0.7818}
            & 27.56 / \textcolor{blue}{0.7359}
            & \textcolor{blue}{26.12} / \textcolor{red}{0.7865} \\
            
            RLFN-S\_ws\_1
            & \textcolor{red}{32.21} / \textcolor{red}{0.8959}
            & \textcolor{red}{28.60} / \textcolor{red}{0.7818}
            & \textcolor{red}{27.57} / \textcolor{red}{0.7360}
            & \textcolor{red}{26.12} / \textcolor{blue}{0.7864} \\
            \bottomrule
        \end{tabular}
    }
    \caption{Effect of learning rate strategy for 4x SR. RLFN-S\_e2000 and RLFN-S\_clr set the total epochs to 2000 to be compared with our proposed strategy RLFN-S\_ws\_1. RLFN-S\_e2000 halves the learning rate every $4\times10^{5}$ iterations. RLFN-S\_clr applies a cyclical learning rate policy. The best and second-best results are marked in \textcolor{red}{red} and \textcolor{blue}{blue} colors, respectively. }
    \label{tab:ab_training_strategy}
    \vspace{-1em}
\end{table}

% \subsubsection{Generalization}
\textbf{Generalization}
We also investigate the generalization of our proposed contrastive loss and warm-start strategy. We apply contrastive loss and warm-start strategy individually to EDSR\cite{EDSR}. The quantitative comparison is shown in Table~\ref{tab:ab_generalization}, which demonstrates that our proposed methods are generic and can be applied to other existing SISR models.

\begin{table}[!htb]
% \vspace{1em}
% \renewcommand{\arraystretch}{1.2}
    \centering
    \scalebox{0.65}{
        \begin{tabular}{c c c c c}
            \toprule
            \multirow{2}{*}{Model} & Set5 & Set14 & BSD100 & Urban100 \\
            ~ & PSNR / SSIM & PSNR / SSIM & PSNR / SSIM & PSNR / SSIM\\
            
            \midrule
            EDSR
            & 32.06 / 0.8945
            & 28.57 / 0.7816
            & 27.56 / 0.7359
            & 26.08 / 0.7859 \\
            \midrule
            
            EDSR + $CL$
            & 32.07 / 0.8946
            & 28.58 / 0.7818
            & 27.57 / 0.7360
            & 26.08 / 0.7861 \\
            
            EDSR\_ws\_1
            & 32.07 / 0.8947
            & 28.59 / 0.7821
            & 27.59 / 0.7363
            & 26.09 / 0.7865
            \\
            \bottomrule
        \end{tabular}
    }
    \caption{Generalization of our proposed contrastive loss and warm-start strategy. We compare EDSR\cite{EDSR} and its two variants which employ the contrastive loss and warm-start strategy, respectively. It can be seen that our proposed methods are generic to existing SISR models.}
    \label{tab:ab_generalization}
    \vspace{-1em}
\end{table}

\subsection{RLFN for NTIRE 2022 challenge}
\begin{table}[t]
% \vspace{1em}
% \renewcommand{\arraystretch}{1.2}
    \centering
    \scalebox{0.5}
    {
        \begin{tabular}{c | c c | c c c c c c}
            \toprule
            \multirow{2}{*}{Team name} & PSNR & PSNR & Ave Time &
            Parameters & FLOPs & Activations & Memery & \multirow{2}{*}{Conv} \\
            ~ & [val] & [test] & [ms] & [M] & [G] & [M] & [M] & ~ \\
            \midrule
            ByteESR(ours) & 29.00 & 28.72 & 27.11 & 0.317 & 19.7 & 80.05 & 377.91 & 39 \\
            
            NJU\_Jet & 29.00 & 28.69 & 28.07 & 0.341 & 22.28 & 72.09 & 204.6 & 34 \\
            
            NEESR & 29.01 & 28.71 & 29.97 & 0.272 & 16.86 & 79.59 & 575.99 & 59 \\
            
            Super & 29.00 & 28.71 & 32.09 & 0.326 & 20.06 & 93.82 & 663.07 & 59 \\
            
            MegSR & 29.00 & 28.68 & 32.59 & 0.29 & 17.7 & 91.72 & 640.63 &64\\
            \midrule
            RFDN(Winner AIM20) & 29.04 & 28.75 & 41.97 & 0.433 & 27.1 & 112.03 & 788.13 & 64 \\
            
            IMDN(Baseline)
            & 29.13 & 28.78 & 50.86 & 0.894
            & 58.53 & 154.14 & 471.76 & 43 \\
            \bottomrule
        \end{tabular}
    }
    \caption{Runtime track results of NTIRE 2022 efficient SR challenge. Only the top five methods are included. }
    \label{challenge_result}
\vspace{-1.5em}
\end{table}

Our team won the 1st place in the main track (Runtime Track) and the 2nd place in the sub-track2 (Overall Performance Track) of NTIRE 2022 efficient super-resolution challenge\cite{li2022ntire}. The model structure and training strategy are slightly different from the above. The proposed RLFN-cut has 4 RLFBs, in which the number of feature channels is set to 48 while the channel number of ESA is set to 16. During training, DIV2K and Flickr2K datasets are used for the whole process.  First, the model is trained from scratch. HR patches of size $256 \times 256$ are randomly cropped from HR images, and the mini-batch size is set to 64.  The model is trained by minimizing L1 loss function with Adam optimizer. The initial learning rate is set to 5e-4 and halved at every 200 epochs.The total number of epochs is 1000.  Then we employ warm-start  policy and train the model with the same settings twice. After that, we change the loss to L1 loss + 255$\times$Contrastive loss, and train with warm-start policy again. At last, we reduce the channels of conv-1 and its dependent conv-3 layers from 48 to 46 using Soft Filter Pruning\cite{SFP}. Training settings 
remain the same except that the size of HR patches changes to $512\times512$. After pruning stage, L2 loss is used for fine-tuning with $640\times640$ HR patches and a learning rate of 1e-5 for 200 epochs. We include the top five methods in Table~\ref{challenge_result}, Compared to baseline IMDN and the first place method RFDN in AIM 2020 Efficient Super-Resolution Challenge, our method achieves significant improvements in all metrics, and we achieve the shortest running time.

% \section{Discussion}

\section{Conclusion}
In this paper, we propose a Residual Local Feature Network for efficient SISR. By reducing the number of layers and simplifying the connections between layers, our network is much lighter and faster. Then we revisit the use of contrastive loss, change the structure of the feature extractor and re-select the intermediate features used by contrastive loss. We also propose a warm-start strategy, which is beneficial on the training of lightweight SR models. Extensive experiments have shown that our overall scheme, including the model structure and training method, achieves a commendable balance of quality and inference speed.

%%%%%%%%% REFERENCES
\clearpage

{\small
\bibliographystyle{ieee_fullname}
\bibliography{egbib}

\begin{thebibliography}{10}\itemsep=-1pt

\bibitem{DIV2K}
Eirikur Agustsson and Radu Timofte.
\newblock Ntire 2017 challenge on single image super-resolution: Dataset and
  study.
\newblock In {\em The IEEE Conference on Computer Vision and Pattern
  Recognition (CVPR) Workshops}, July 2017.

\bibitem{CARN}
Namhyuk Ahn, Byungkon Kang, and Kyung{-}Ah Sohn.
\newblock Fast, accurate, and lightweight super-resolution with cascading
  residual network.
\newblock In Vittorio Ferrari, Martial Hebert, Cristian Sminchisescu, and Yair
  Weiss, editors, {\em Computer Vision - {ECCV} 2018 - 15th European
  Conference, Munich, Germany, September 8-14, 2018, Proceedings, Part {X}},
  volume 11214 of {\em Lecture Notes in Computer Science}, pages 256--272.
  Springer, 2018.

\bibitem{revisitingresnet}
Irwan Bello, William Fedus, Xianzhi Du, Ekin~Dogus Cubuk, Aravind Srinivas,
  Tsung-Yi Lin, Jonathon Shlens, and Barret Zoph.
\newblock Revisiting resnets: Improved training and scaling strategies.
\newblock {\em Advances in Neural Information Processing Systems}, 34, 2021.

\bibitem{Set5}
Marco Bevilacqua, Aline Roumy, Christine Guillemot, and Marie~Line
  Alberi-Morel.
\newblock Low-complexity single-image super-resolution based on nonnegative
  neighbor embedding.
\newblock 2012.

\bibitem{IPT}
Hanting Chen, Yunhe Wang, Tianyu Guo, Chang Xu, Yiping Deng, Zhenhua Liu, Siwei
  Ma, Chunjing Xu, Chao Xu, and Wen Gao.
\newblock Pre-trained image processing transformer.
\newblock In {\em {IEEE} Conference on Computer Vision and Pattern Recognition,
  {CVPR} 2021, virtual, June 19-25, 2021}, pages 12299--12310. Computer Vision
  Foundation / {IEEE}, 2021.

\bibitem{SimCLR}
Ting Chen, Simon Kornblith, Mohammad Norouzi, and Geoffrey Hinton.
\newblock A simple framework for contrastive learning of visual
  representations.
\newblock In {\em International conference on machine learning}, pages
  1597--1607. PMLR, 2020.

\bibitem{SimCLRV2}
Ting Chen, Simon Kornblith, Kevin Swersky, Mohammad Norouzi, and Geoffrey~E.
  Hinton.
\newblock Big self-supervised models are strong semi-supervised learners.
\newblock In Hugo Larochelle, Marc'Aurelio Ranzato, Raia Hadsell,
  Maria{-}Florina Balcan, and Hsuan{-}Tien Lin, editors, {\em Advances in
  Neural Information Processing Systems 33: Annual Conference on Neural
  Information Processing Systems 2020, NeurIPS 2020, December 6-12, 2020,
  virtual}, 2020.

\bibitem{MoCoV2}
Xinlei Chen, Haoqi Fan, Ross~B. Girshick, and Kaiming He.
\newblock Improved baselines with momentum contrastive learning.
\newblock {\em CoRR}, abs/2003.04297, 2020.

\bibitem{FSRNet}
Yu Chen, Ying Tai, Xiaoming Liu, Chunhua Shen, and Jian Yang.
\newblock Fsrnet: End-to-end learning face super-resolution with facial priors.
\newblock In {\em 2018 {IEEE} Conference on Computer Vision and Pattern
  Recognition, {CVPR} 2018, Salt Lake City, UT, USA, June 18-22, 2018}, pages
  2492--2501. Computer Vision Foundation / {IEEE} Computer Society, 2018.

\bibitem{REPVGG}
Xiaohan Ding, Xiangyu Zhang, Ningning Ma, Jungong Han, Guiguang Ding, and Jian
  Sun.
\newblock Repvgg: Making vgg-style convnets great again.
\newblock In {\em Proceedings of the IEEE/CVF Conference on Computer Vision and
  Pattern Recognition}, pages 13733--13742, 2021.

\bibitem{SRCNN}
Chao Dong, Chen~Change Loy, Kaiming He, and Xiaoou Tang.
\newblock Learning a deep convolutional network for image super-resolution.
\newblock In David~J. Fleet, Tom{\'{a}}s Pajdla, Bernt Schiele, and Tinne
  Tuytelaars, editors, {\em Computer Vision - {ECCV} 2014 - 13th European
  Conference, Zurich, Switzerland, September 6-12, 2014, Proceedings, Part
  {IV}}, volume 8692 of {\em Lecture Notes in Computer Science}, pages
  184--199. Springer, 2014.

\bibitem{FSRCNN}
Chao Dong, Chen~Change Loy, and Xiaoou Tang.
\newblock Accelerating the super-resolution convolutional neural network.
\newblock In Bastian Leibe, Jiri Matas, Nicu Sebe, and Max Welling, editors,
  {\em Computer Vision - {ECCV} 2016 - 14th European Conference, Amsterdam, The
  Netherlands, October 11-14, 2016, Proceedings, Part {II}}, volume 9906 of
  {\em Lecture Notes in Computer Science}, pages 391--407. Springer, 2016.

\bibitem{HyperspectralSR}
Ying Fu, Tao Zhang, Yinqiang Zheng, Debing Zhang, and Hua Huang.
\newblock Hyperspectral image super-resolution with optimized {RGB} guidance.
\newblock In {\em {IEEE} Conference on Computer Vision and Pattern Recognition,
  {CVPR} 2019, Long Beach, CA, USA, June 16-20, 2019}, pages 11661--11670.
  Computer Vision Foundation / {IEEE}, 2019.

\bibitem{FDIWN}
Guangwei Gao, Wenjie Li, Juncheng Li, Fei Wu, Huimin Lu, and Yi Yu.
\newblock Feature distillation interaction weighting network for lightweight
  image super-resolution.
\newblock {\em CoRR}, abs/2112.08655, 2021.

\bibitem{MoCo}
Kaiming He, Haoqi Fan, Yuxin Wu, Saining Xie, and Ross Girshick.
\newblock Momentum contrast for unsupervised visual representation learning.
\newblock In {\em Proceedings of the IEEE/CVF Conference on Computer Vision and
  Pattern Recognition}, pages 9729--9738, 2020.

\bibitem{SFP}
Yang He, Guoliang Kang, Xuanyi Dong, Yanwei Fu, and Yi Yang.
\newblock Soft filter pruning for accelerating deep convolutional neural
  networks.
\newblock {\em arXiv preprint arXiv:1808.06866}, 2018.

\bibitem{Urban100}
Jia-Bin Huang, Abhishek Singh, and Narendra Ahuja.
\newblock Single image super-resolution from transformed self-exemplars.
\newblock In {\em Proceedings of the IEEE conference on computer vision and
  pattern recognition}, pages 5197--5206, 2015.

\bibitem{IMDN}
Zheng Hui, Xinbo Gao, Yunchu Yang, and Xiumei Wang.
\newblock Lightweight image super-resolution with information
  multi-distillation network.
\newblock In Laurent Amsaleg, Benoit Huet, Martha~A. Larson, Guillaume Gravier,
  Hayley Hung, Chong{-}Wah Ngo, and Wei~Tsang Ooi, editors, {\em Proceedings of
  the 27th {ACM} International Conference on Multimedia, {MM} 2019, Nice,
  France, October 21-25, 2019}, pages 2024--2032. {ACM}, 2019.

\bibitem{IDN}
Zheng Hui, Xiumei Wang, and Xinbo Gao.
\newblock Fast and accurate single image super-resolution via information
  distillation network.
\newblock In {\em Proceedings of the IEEE conference on computer vision and
  pattern recognition}, pages 723--731, 2018.

\bibitem{MAI2021SR}
Andrey Ignatov, Radu Timofte, Maurizio Denna, and Abdel Younes.
\newblock Real-time quantized image super-resolution on mobile npus, mobile
  {AI} 2021 challenge: Report.
\newblock In {\em {IEEE} Conference on Computer Vision and Pattern Recognition
  Workshops, {CVPR} Workshops 2021, virtual, June 19-25, 2021}, pages
  2525--2534. Computer Vision Foundation / {IEEE}, 2021.

\bibitem{CLSurvey}
Ashish Jaiswal, Ashwin~Ramesh Babu, Mohammad~Zaki Zadeh, Debapriya Banerjee,
  and Fillia Makedon.
\newblock A survey on contrastive self-supervised learning.
\newblock {\em Technologies}, 9(1):2, 2021.

\bibitem{VDSR}
Jiwon Kim, Jung~Kwon Lee, and Kyoung~Mu Lee.
\newblock Accurate image super-resolution using very deep convolutional
  networks.
\newblock In {\em 2016 {IEEE} Conference on Computer Vision and Pattern
  Recognition, {CVPR} 2016, Las Vegas, NV, USA, June 27-30, 2016}, pages
  1646--1654. {IEEE} Computer Society, 2016.

\bibitem{DRCN}
Jiwon Kim, Jung~Kwon Lee, and Kyoung~Mu Lee.
\newblock Deeply-recursive convolutional network for image super-resolution.
\newblock In {\em 2016 {IEEE} Conference on Computer Vision and Pattern
  Recognition, {CVPR} 2016, Las Vegas, NV, USA, June 27-30, 2016}, pages
  1637--1645. {IEEE} Computer Society, 2016.

\bibitem{Adam}
Diederik~P. Kingma and Jimmy Ba.
\newblock Adam: {A} method for stochastic optimization.
\newblock In Yoshua Bengio and Yann LeCun, editors, {\em 3rd International
  Conference on Learning Representations, {ICLR} 2015, San Diego, CA, USA, May
  7-9, 2015, Conference Track Proceedings}, 2015.

\bibitem{LapSRN}
Wei{-}Sheng Lai, Jia{-}Bin Huang, Narendra Ahuja, and Ming{-}Hsuan Yang.
\newblock Deep laplacian pyramid networks for fast and accurate
  super-resolution.
\newblock In {\em 2017 {IEEE} Conference on Computer Vision and Pattern
  Recognition, {CVPR} 2017, Honolulu, HI, USA, July 21-26, 2017}, pages
  5835--5843. {IEEE} Computer Society, 2017.

\bibitem{FP}
Hao Li, Asim Kadav, Igor Durdanovic, Hanan Samet, and Hans~Peter Graf.
\newblock Pruning filters for efficient convnets.
\newblock {\em arXiv preprint arXiv:1608.08710}, 2016.

\bibitem{li2022ntire}
Yawei Li, Kai Zhang, Luc~Van Gool, Radu Timofte, et~al.
\newblock Ntire 2022 challenge on efficient super-resolution: Methods and
  results.
\newblock In {\em IEEE Conference on Computer Vision and Pattern Recognition
  Workshops}, 2022.

\bibitem{SwinIR}
Jingyun Liang, Jiezhang Cao, Guolei Sun, Kai Zhang, Luc~Van Gool, and Radu
  Timofte.
\newblock Swinir: Image restoration using swin transformer.
\newblock In {\em {IEEE/CVF} International Conference on Computer Vision
  Workshops, {ICCVW} 2021, Montreal, BC, Canada, October 11-17, 2021}, pages
  1833--1844. {IEEE}, 2021.

\bibitem{EDSR}
Bee Lim, Sanghyun Son, Heewon Kim, Seungjun Nah, and Kyoung~Mu Lee.
\newblock Enhanced deep residual networks for single image super-resolution.
\newblock In {\em 2017 {IEEE} Conference on Computer Vision and Pattern
  Recognition Workshops, {CVPR} Workshops 2017, Honolulu, HI, USA, July 21-26,
  2017}, pages 1132--1140. {IEEE} Computer Society, 2017.

\bibitem{lin2022revisiting}
Zudi Lin, Prateek Garg, Atmadeep Banerjee, Salma~Abdel Magid, Deqing Sun, Yulun
  Zhang, Luc Van~Gool, Donglai Wei, and Hanspeter Pfister.
\newblock Revisiting rcan: Improved training for image super-resolution.
\newblock {\em arXiv preprint arXiv:2201.11279}, 2022.

\bibitem{RFDN}
Jie Liu, Jie Tang, and Gangshan Wu.
\newblock Residual feature distillation network for lightweight image
  super-resolution.
\newblock In Adrien Bartoli and Andrea Fusiello, editors, {\em Computer Vision
  - {ECCV} 2020 Workshops - Glasgow, UK, August 23-28, 2020, Proceedings, Part
  {III}}, volume 12537 of {\em Lecture Notes in Computer Science}, pages
  41--55. Springer, 2020.

\bibitem{RFANet}
Jie Liu, Wenjie Zhang, Yuting Tang, Jie Tang, and Gangshan Wu.
\newblock Residual feature aggregation network for image super-resolution.
\newblock In {\em Proceedings of the IEEE/CVF conference on computer vision and
  pattern recognition}, pages 2359--2368, 2020.

\bibitem{GenericPerceptualLoss}
Yifan Liu, Hao Chen, Yu Chen, Wei Yin, and Chunhua Shen.
\newblock Generic perceptual loss for modeling structured output dependencies.
\newblock In {\em Proceedings of the IEEE/CVF Conference on Computer Vision and
  Pattern Recognition (CVPR)}, pages 5424--5432, June 2021.

\bibitem{DFSA}
Salma~Abdel Magid, Yulun Zhang, Donglai Wei, Won{-}Dong Jang, Zudi Lin, Yun Fu,
  and Hanspeter Pfister.
\newblock Dynamic high-pass filtering and multi-spectral attention for image
  super-resolution.
\newblock In {\em 2021 {IEEE/CVF} International Conference on Computer Vision,
  {ICCV} 2021, Montreal, QC, Canada, October 10-17, 2021}, pages 4268--4277.
  {IEEE}, 2021.

\bibitem{BSD100}
David Martin, Charless Fowlkes, Doron Tal, and Jitendra Malik.
\newblock A database of human segmented natural images and its application to
  evaluating segmentation algorithms and measuring ecological statistics.
\newblock In {\em Proceedings Eighth IEEE International Conference on Computer
  Vision. ICCV 2001}, volume~2, pages 416--423. IEEE, 2001.

\bibitem{NLSA}
Yiqun Mei, Yuchen Fan, and Yuqian Zhou.
\newblock Image super-resolution with non-local sparse attention.
\newblock In {\em {IEEE} Conference on Computer Vision and Pattern Recognition,
  {CVPR} 2021, virtual, June 19-25, 2021}, pages 3517--3526. Computer Vision
  Foundation / {IEEE}, 2021.

\bibitem{MAFFSRN}
Abdul Muqeet, Jiwon Hwang, Subin Yang, Jung~Heum Kang, Yongwoo Kim, and
  Sung{-}Ho Bae.
\newblock Multi-attention based ultra lightweight image super-resolution.
\newblock In Adrien Bartoli and Andrea Fusiello, editors, {\em Computer Vision
  - {ECCV} 2020 Workshops - Glasgow, UK, August 23-28, 2020, Proceedings, Part
  {III}}, volume 12537 of {\em Lecture Notes in Computer Science}, pages
  103--118. Springer, 2020.

\bibitem{HAN}
Ben Niu, Weilei Wen, Wenqi Ren, Xiangde Zhang, Lianping Yang, Shuzhen Wang,
  Kaihao Zhang, Xiaochun Cao, and Haifeng Shen.
\newblock Single image super-resolution via a holistic attention network.
\newblock In Andrea Vedaldi, Horst Bischof, Thomas Brox, and Jan{-}Michael
  Frahm, editors, {\em Computer Vision - {ECCV} 2020 - 16th European
  Conference, Glasgow, UK, August 23-28, 2020, Proceedings, Part {XII}}, volume
  12357 of {\em Lecture Notes in Computer Science}, pages 191--207. Springer,
  2020.

\bibitem{ESPCN}
Wenzhe Shi, Jose Caballero, Ferenc Huszar, Johannes Totz, Andrew~P. Aitken, Rob
  Bishop, Daniel Rueckert, and Zehan Wang.
\newblock Real-time single image and video super-resolution using an efficient
  sub-pixel convolutional neural network.
\newblock In {\em 2016 {IEEE} Conference on Computer Vision and Pattern
  Recognition, {CVPR} 2016, Las Vegas, NV, USA, June 27-30, 2016}, pages
  1874--1883. {IEEE} Computer Society, 2016.

\bibitem{DepthMapSR}
Xibin Song, Yuchao Dai, Dingfu Zhou, Liu Liu, Wei Li, Hongdong Li, and Ruigang
  Yang.
\newblock Channel attention based iterative residual learning for depth map
  super-resolution.
\newblock In {\em 2020 {IEEE/CVF} Conference on Computer Vision and Pattern
  Recognition, {CVPR} 2020, Seattle, WA, USA, June 13-19, 2020}, pages
  5630--5639. Computer Vision Foundation / {IEEE}, 2020.

\bibitem{sun2021autoflow}
Deqing Sun, Daniel Vlasic, Charles Herrmann, Varun Jampani, Michael Krainin,
  Huiwen Chang, Ramin Zabih, William~T Freeman, and Ce Liu.
\newblock Autoflow: Learning a better training set for optical flow.
\newblock In {\em Proceedings of the IEEE/CVF Conference on Computer Vision and
  Pattern Recognition}, pages 10093--10102, 2021.

\bibitem{DRRN}
Ying Tai, Jian Yang, and Xiaoming Liu.
\newblock Image super-resolution via deep recursive residual network.
\newblock In {\em 2017 {IEEE} Conference on Computer Vision and Pattern
  Recognition, {CVPR} 2017, Honolulu, HI, USA, July 21-26, 2017}, pages
  2790--2798. {IEEE} Computer Society, 2017.

\bibitem{RandomStyleTransfer}
Pei Wang, Yijun Li, and Nuno Vasconcelos.
\newblock Rethinking and improving the robustness of image style transfer.
\newblock In {\em Proceedings of the IEEE/CVF Conference on Computer Vision and
  Pattern Recognition (CVPR)}, pages 124--133, June 2021.

\bibitem{CSD}
Yanbo Wang, Shaohui Lin, Yanyun Qu, Haiyan Wu, Zhizhong Zhang, Yuan Xie, and
  Angela Yao.
\newblock Towards compact single image super-resolution via contrastive
  self-distillation.
\newblock In Zhi{-}Hua Zhou, editor, {\em Proceedings of the Thirtieth
  International Joint Conference on Artificial Intelligence, {IJCAI} 2021,
  Virtual Event / Montreal, Canada, 19-27 August 2021}, pages 1122--1128.
  ijcai.org, 2021.

\bibitem{AIM2020}
Pengxu Wei, Hannan Lu, Radu Timofte, Liang Lin, Wangmeng Zuo, Zhihong Pan,
  Baopu Li, Teng Xi, Yanwen Fan, Gang Zhang, Jingtuo Liu, Junyu Han, Errui
  Ding, Tangxin Xie, Liang Cao, Yan Zou, Yi Shen, Jialiang Zhang, Yu Jia,
  Kaihua Cheng, Chenhuan Wu, Yue Lin, Cen Liu, Yunbo Peng, Xueyi Zou, Zhipeng
  Luo, Yuehan Yao, Zhenyu Xu, Syed~Waqas Zamir, Aditya Arora, Salman~H. Khan,
  Munawar Hayat, Fahad~Shahbaz Khan, Keon{-}Hee Ahn, Jun{-}Hyuk Kim, Jun{-}Ho
  Choi, Jong{-}Seok Lee, Tongtong Zhao, Shanshan Zhao, Yoseob Han, Byung{-}Hoon
  Kim, JaeHyun Baek, Haoning Wu, Dejia Xu, Bo Zhou, Wei Guan, Xiaobo Li, Chen
  Ye, Hao Li, Haoyu Zhong, Yukai Shi, Zhijing Yang, Xiaojun Yang, Xin Li, Xin
  Jin, Yaojun Wu, Yingxue Pang, Sen Liu, Zhi{-}Song Liu, Li{-}Wen Wang,
  Chu{-}Tak Li, Marie{-}Paule Cani, Wan{-}Chi Siu, Yuanbo Zhou, Rao~Muhammad
  Umer, Christian Micheloni, Xiaofeng Cong, Rajat Gupta, Feras Almasri, Thomas
  Vandamme, and Olivier Debeir.
\newblock {AIM} 2020 challenge on real image super-resolution: Methods and
  results.
\newblock In Adrien Bartoli and Andrea Fusiello, editors, {\em Computer Vision
  - {ECCV} 2020 Workshops - Glasgow, UK, August 23-28, 2020, Proceedings, Part
  {III}}, volume 12537 of {\em Lecture Notes in Computer Science}, pages
  392--422. Springer, 2020.

\bibitem{AECR-Net}
Haiyan Wu, Yanyun Qu, Shaohui Lin, Jian Zhou, Ruizhi Qiao, Zhizhong Zhang, Yuan
  Xie, and Lizhuang Ma.
\newblock Contrastive learning for compact single image dehazing.
\newblock In {\em {IEEE} Conference on Computer Vision and Pattern Recognition,
  {CVPR} 2021, virtual, June 19-25, 2021}, pages 10551--10560. Computer Vision
  Foundation / {IEEE}, 2021.

\bibitem{Set14}
Roman Zeyde, Michael Elad, and Matan Protter.
\newblock On single image scale-up using sparse-representations.
\newblock In {\em International conference on curves and surfaces}, pages
  711--730. Springer, 2010.

\bibitem{ECBSR}
Xindong Zhang, Hui Zeng, and Lei Zhang.
\newblock Edge-oriented convolution block for real-time super resolution on
  mobile devices.
\newblock In Heng~Tao Shen, Yueting Zhuang, John~R. Smith, Yang Yang, Pablo
  Cesar, Florian Metze, and Balakrishnan Prabhakaran, editors, {\em {MM} '21:
  {ACM} Multimedia Conference, Virtual Event, China, October 20 - 24, 2021},
  pages 4034--4043. {ACM}, 2021.

\bibitem{RCAN}
Yulun Zhang, Kunpeng Li, Kai Li, Lichen Wang, Bineng Zhong, and Yun Fu.
\newblock Image super-resolution using very deep residual channel attention
  networks.
\newblock In Vittorio Ferrari, Martial Hebert, Cristian Sminchisescu, and Yair
  Weiss, editors, {\em Computer Vision - {ECCV} 2018 - 15th European
  Conference, Munich, Germany, September 8-14, 2018, Proceedings, Part {VII}},
  volume 11211 of {\em Lecture Notes in Computer Science}, pages 294--310.
  Springer, 2018.

\end{thebibliography}
}

\end{document}